# Monte Carlo Methods for Tempo Tracking and Rhythm Quantization

**Ali Taylan Cemgil** CEMGIL@SNN.KUN.NL
**Bert Kappen** BERT@SNN.KUN.NL
*SNN, Geert Grooteplein 21 cpk1 - 231, University of Nijmegen*
*NL 6525 EZ Nijmegen, The Netherlands*

## Abstract

We present a probabilistic generative model for timing deviations in expressive music performance. The structure of the proposed model is equivalent to a switching state space model. The switch variables correspond to discrete note locations as in a musical score. The continuous hidden variables denote the tempo. We formulate two well known music recognition problems, namely tempo tracking and automatic transcription (rhythm quantization) as filtering and maximum a posteriori (MAP) state estimation tasks. Exact computation of posterior features such as the MAP state is intractable in this model class, so we introduce Monte Carlo methods for integration and optimization. We compare Markov Chain Monte Carlo (MCMC) methods (such as Gibbs sampling, simulated annealing and iterative improvement) and sequential Monte Carlo methods (particle filters). Our simulation results suggest better results with sequential methods. The methods can be applied in both online and batch scenarios such as tempo tracking and transcription and are thus potentially useful in a number of music applications such as adaptive automatic accompaniment, score typesetting and music information retrieval.

## 1. Introduction

Automatic music transcription refers to extraction of a human readable and interpretable description from a recording of a musical performance. Traditional music notation is such a description that lists the pitch levels (notes) and corresponding timestamps.

Ideally, one would like to recover a score directly from the audio signal. Such a representation of the surface structure of music would be very useful in music information retrieval (Music-IR) and content description of musical material in large audio databases. However, when operating on sampled audio data from polyphonic acoustical signals, extraction of a score-like description is a very challenging auditory scene analysis task (Vercoe, Gardner, & Scheirer, 1998).

In this paper, we focus on a subproblem in music-ir, where we assume that exact timing information of notes is available, for example as a stream of MIDI[1] events from a digital keyboard.

A model for tempo tracking and transcription from a MIDI-like music representation is useful in a broad spectrum of applications. One example is automatic score typesetting,

1. Musical Instruments Digital Interface. A standard communication protocol especially designed for digital instruments such as keyboards. Each time a key is pressed, a MIDI keyboard generates a short message containing pitch and key velocity. A computer can tag each received message by a timestamp for real-time processing and/or recording into a file.

the musical analog of word processing. Almost all score typesetting applications provide a means of automatic generation of a conventional music notation from MIDI data.

In conventional music notation, the onset time of each note is implicitly represented by the cumulative sum of durations of previous notes. Durations are encoded by simple rational numbers (e.g., quarter note, eighth note), consequently all events in music are placed on a discrete grid. So the basic task in MIDI transcription is to associate onset times with discrete grid locations, i.e., quantization.

However, unless the music is performed with mechanical precision, identification of the correct association becomes difficult. This is due to the fact that musicians introduce intentional (and unintentional) deviations from a mechanical prescription. For example timing of events can be deliberately delayed or pushed. Moreover, the tempo can fluctuate by slowing down or accelerating. In fact, such deviations are natural aspects of expressive performance; in the absence of these, music tends to sound rather dull and mechanical. On the other hand, if these deviations are not accounted for during transcription, resulting scores have often very poor quality.

Robust and fast quantization and tempo tracking is also an important requirement for interactive performance systems; applications that "listen" to a performer for generating an accompaniment or improvisation in real time (Raphael, 2001b; Thom, 2000). At last, such models are also useful in musicology for systematic study and characterization of expressive timing by principled analysis of existing performance data.

From a theoretical perspective, simultaneous quantization *and* tempo tracking is a "chicken-and-egg" problem: the quantization depends upon the intended tempo interpretation and the tempo interpretation depends upon the quantization. Apparently, human listeners can resolve this ambiguity (in most cases) without any effort. Even persons without any musical training are able to determine the beat and the tempo very rapidly. However, it is still unclear what precisely constitutes tempo and how it relates to the perception of the beat, rhythmical structure, pitch, style of music etc. Tempo is a perceptual construct and cannot directly be measured in a performance.

The goal of understanding tempo perception has stimulated a significant body of research on the psychological and computational modeling aspects of tempo tracking and beat induction, e.g., see (Desain & Honing, 1994; Large & Jones, 1999; Toiviainen, 1999). These papers assume that events are presented as an onset list. Attempts are also made to deal directly with the audio signal (Goto & Muraoka, 1998; Scheirer, 1998; Dixon & Cambouropoulos, 2000).

Another class of tempo tracking models are developed in the context of interactive performance systems and score following. These models make use of prior knowledge in the form of an annotated score (Dannenberg, 1984; Vercoe & Puckette, 1985). More recently, Raphael (2001b) has demonstrated an interactive real-time system that follows a solo player and schedules accompaniment events according to the player's tempo interpretation.

Tempo tracking is crucial for quantization, since one can not uniquely quantize onsets without having an estimate of tempo and the beat. The converse, that quantization can help in identification of the correct tempo interpretation has already been noted by Desain and Honing (1991). Here, one defines correct tempo as the one that results in a simpler quantization. However, such a schema has never been fully implemented in practice due to computational complexity of obtaining a perceptually plausible quantization. Hence quantization methods proposed in the literature either estimate the tempo using simple heuristics (Longuet-Higgins, 1987; Pressing & Lawrence, 1993; Agon, Assayag, Fineberg, & Rueda, 1994) or assume that the tempo is known or constant (Desain & Honing, 1991; Cambouropoulos, 2000; Hamanaka, Goto, Asoh, & Otsu, 2001).

Our approach to transcription and tempo tracking is from a probabilistic, i.e., Bayesian modeling perspective. In Cemgil et al. (2000), we introduced a probabilistic approach to perceptually realistic quantization. This work also assumed that the tempo was known or was estimated by an external procedure. For tempo tracking, we introduced a Kalman filter model (Cemgil, Kappen, Desain, & Honing, 2001). In this approach, we modeled the tempo as a smoothly varying hidden state variable of a stochastic dynamical system.

In the current paper, we integrate quantization and tempo tracking. Basically, our model balances score complexity versus smoothness in tempo deviations. The correct tempo interpretation results in a simple quantization and the correct quantization results in a smooth tempo fluctuation. An essentially similar model is proposed recently also by Raphael (2001a). However, Raphael uses an inference technique that only applies for small models; namely when the continuous hidden state is one dimensional. This severely restricts the models one can consider. In the current paper, we survey general and widely used state-of-the-art techniques for inference.

The outline of the paper is as follows: In Section 2, we propose a probabilistic model for timing deviations in expressive music performance. Given the model, we will define tempo tracking and quantization as inference of posterior quantities. It will turn out that our model is a switching state space model in which computation of exact probabilities becomes intractable. In Section 3, we will introduce approximation techniques based on simulation, namely Markov Chain Monte Carlo (MCMC) and sequential Monte Carlo (SMC) (Doucet, de Freitas, & Gordon, 2001; Andrieu, de Freitas, Doucet, & Jordan, 2002). Both approaches provide flexible and powerful inference methods that have been successfully applied in diverse fields of applied sciences such as robotics (Fox, Burgard, & Thrun, 1999), aircraft tracking (Gordon, Salmond, & Smith, 1993), computer vision (Isard & Blake, 1996), econometrics (Tanizaki, 2001). Finally we will present simulation results and conclusions.

## 2. Model

Assume that a pianist is improvising and we are recording the exact onset times of each key she presses during the performance. We denote these observed onset times by $y_0, y_1, y_2 \ldots y_k \ldots y_K$ or more compactly by $y_{0:K}$. We neither have access to a musical notation of the piece nor know the initial tempo she has started her performance with. Moreover, the pianist is allowed to freely change the tempo or introduce expression. Given only onset time information $y_{0:K}$, we wish to find a score $\gamma_{1:K}$ and track her tempo fluctuations $z_{0:K}$. We will refine the meaning of $\gamma$ and $z$ later.

This problem is apparently ill-posed. If the pianist is allowed to change the tempo arbitrarily it is not possible to assign a "correct" score to a given performance. In other words any performance $y_{0:K}$ can be represented by using a suitable combination of an arbitrary score with an arbitrary tempo trajectory. Fortunately, the Bayes theorem provides an elegant and principled guideline to formulate the problem. Given the onsets $y_{0:K}$, the best score $\gamma_{1:K}$ and tempo trajectory $z_{0:K}$ can be derived from the *posterior* distribution

that is given by

$$p(\gamma_{1:K}, z_{0:K}|y_{0:K}) \;=\; \frac{1}{p(y_{0:K})} p(y_{0:K}|\gamma_{1:K}, z_{0:K}) p(\gamma_{1:K}, z_{0:K})$$

a quantity, that is proportional to the product of the *likelihood* term $p(y_{0:K}|\gamma_{1:K}, z_{0:K})$ and the *prior* term $p(\gamma_{1:K}, z_{0:K})$.

In rhythm transcription and tempo tracking, the prior encodes our background knowledge about the nature of musical scores and tempo deviations. For example, we can construct a prior that prefers "simple" scores and smooth tempo variations.

The likelihood term relates the tempo and the score to actual observed onset times. In this respect, the likelihood is a model for short time expressive timing deviations and motor errors that are introduced by the performer.

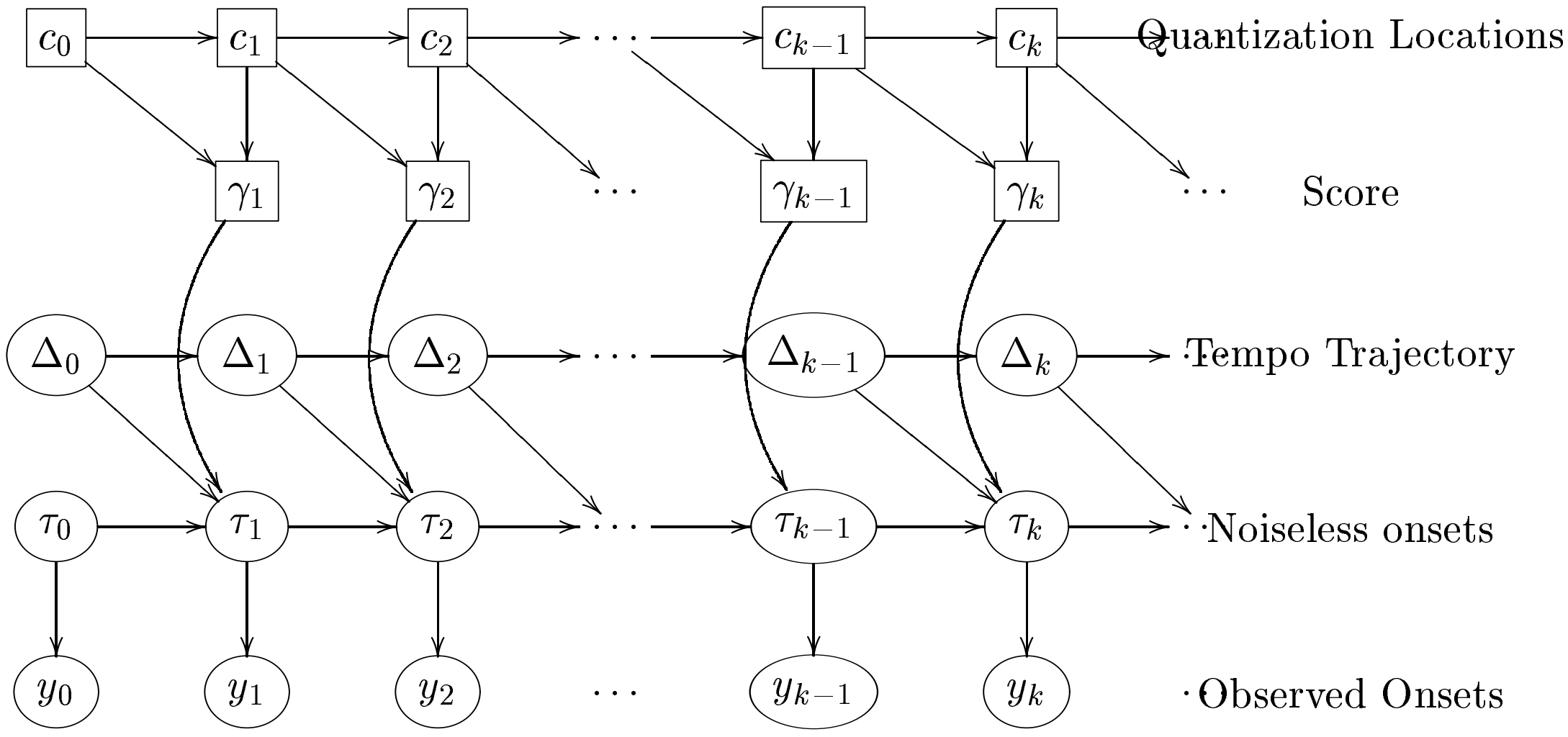

Figure 1: Graphical Model. Square and oval nodes correspond to discrete and continuous variables respectively. In the text, we sometimes refer to the continuous hidden variables $(\tau_k, \Delta_k)$ by $z_k$. The dependence between $\gamma$ and $c$ is deterministic. All $c$, $\gamma$ , $\tau$ and $\Delta$ are hidden; only onsets $y$ are observed.

### 2.1 Score prior

To define a score $\gamma_{1:K}$, we first introduce a sequence of *quantization locations* $c_{0:K}$. A quantization location $c_k$ specifies the score time of the $k$'th onset. We let $\gamma_k$ denote the interval between quantization locations of two consecutive onsets

$$\gamma_k \;=\; c_k - c_{k-1} \tag{1}$$

For example consider the conventional music notation ♩ ♫ which encodes the score $\gamma_{1:3} = [1 \;\; 0.5 \;\; 0.5]$. Corresponding quantization locations are $c_{0:3} = [0 \;\; 1 \;\; 1.5 \;\; 2]$.

One simple way of defining a prior distribution on quantization locations $p(c_k)$ is specifying a table of probabilities for $c_k \mod 1$ (the fraction of $c_k$). For example if we wish to allow for scores that have sixteenth notes and triplets, we define a table of probabilities for the states $c \mod 1 = \{0,\ 0.25,\ 0.5,\ 0.75\} \cup \{0,\ 0.33,\ 0.67\}$. Technically, the resulting prior $p(c_k)$ is periodic and improper (since $c_k$ are in principle unbounded so we can not normalize the distribution).

However, if the number of states of $c_k \mod 1$ is large, it may be difficult to estimate the parameters of the prior reliably. For such situations we propose a "generic" prior as follows: We define the probability, that the $k$'th onset gets quantized at location $c_k$, by $p(c_k) \propto \exp(-\lambda d(c_k))$ where $d(c_k)$ is the number of significant digits in the *binary* expansion of $c_k \mod 1$. For example $d(1) = 0$, $d(1.5) = 1$, $d(7 + 9/32) = 5$ etc. The positive parameter $\lambda$ is used to penalize quantization locations that require more bits to be represented. Assuming that quantization locations of onsets are independent a-priori, (besides being increasing in $k$, i.e., $c_k \geq c_{k-1}$), the prior probability of a sequence of quantization locations is given by $p(c_{0:K}) \propto \exp(-\lambda \sum_{k=0}^{K} d(c_k))$. We further assume that $c_0 \in [0, 1)$. One can check that such a prior prefers simpler notations, e.g., $p($ 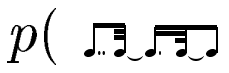 $) < p($ 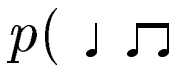 $)$. We can generalize this prior to other subdivisions such triplets and quintiplets in Appendix A.

Formally, given a distribution on $c_{0:K}$, the prior of a score $\gamma_{1:K}$ is given by

$$p(\gamma_{1:K}) = \sum_{c_{0:K}} p(\gamma_{1:K}|c_{0:K}) p(c_{0:K}) \tag{2}$$

Since the relationship between $c_{0:K}$ and $\gamma_{1:K}$ is deterministic, $p(\gamma_{1:K}|c_{0:K})$ is degenerate for any given $c_{0:K}$, so we have

$$p(\gamma_{1:K}) \propto \exp\left(-\lambda \sum_{k=1}^{K} d\left(\sum_{k'=1}^{k} \gamma_{k'}\right)\right) \tag{3}$$

One might be tempted to specify a prior directly on $\gamma_{1:K}$ and get rid of $c_{0:K}$ entirely. However, with this simpler approach it is not easy to devise realistic priors. For example, consider a sequence of note durations $[1\ \ 1/16\ \ 1\ \ 1\ \ 1\ldots]$. Assuming a factorized prior on $\gamma$ that penalizes short note durations, this rhythm would have relatively high probability whereas it is quite uncommon in conventional music.

### 2.2 Tempo prior

We represent the tempo in terms of its inverse, i.e., the period, and denote it with $\Delta$. For example a tempo of 120 beats per minute (bpm) corresponds to $\Delta = 60/120 = 0.5$ seconds. At each onset the tempo changes by an unknown amount $\zeta_{\Delta_k}$. We assume the change $\zeta_{\Delta_k}$ is iid with $\mathcal{N}(0, Q_\Delta)$. [2] We assume a first order Gauss-Markov process for the tempo

$$\Delta_k = \Delta_{k-1} + \zeta_{\Delta_k} \tag{4}$$

Eq. 4 defines a distribution over tempo sequences $\Delta_{0:K}$. Given a tempo sequence, the "ideal" or "intended" time $\tau_k$ of the next onset is given by

$$\tau_k = \tau_{k-1} + \gamma_k \Delta_{k-1} + \zeta_{\tau_k} \tag{5}$$

2. We denote a (scalar or multivariate) Gaussian distribution $p(\mathbf{x})$ with mean vector $\mu$ and covariance matrix $P$ by $\mathcal{N}(\mu, P) \hat{=} |2\pi P|^{-\frac{1}{2}} \exp(-\frac{1}{2}(\mathbf{x} - \mu)^T P^{-1} (\mathbf{x} - \mu))$.

The noise term $\zeta_{\tau_k}$ denotes the amount of accentuation (that is deliberately playing a note ahead or back in time) without causing the tempo to be changed. We assume $\zeta_{\tau_k} \sim \mathcal{N}(0, Q_\tau)$. Ideal onsets and actually observed "noisy" onsets are related by

$$y_k = \tau_k + \epsilon_k \tag{6}$$

The noise term $\epsilon_k$ models small scale expressive deviations or motor errors in timing of individual notes. In this paper we will assume that $\epsilon_k$ has a Gaussian distribution parameterized by $\mathcal{N}(0, R)$.

The initial tempo distribution $p(\Delta_0)$ specifies a range of reasonable tempi and is given by a Gaussian with a broad variance. We assume an uninformative (flat) prior on $\tau_0$. The conditional independence structure is given by the graphical model in Figure 1. Table 1 shows a possible realization from the model.

We note that our model is a particular instance of the well known switching state space model (also known as conditionally linear dynamical system, jump Markov linear system, switching Kalman filter) (See, e.g., Bar-Shalom & Li, 1993; Doucet & Andrieu, 2001; Murphy, 2002).

| $k$ | 0 | 1 | 2 | 3 | |
|---|---|---|---|---|---|
| $\gamma_k$ | | ♪ | ♩ | ♪ | ... |
| $c_k$ | 0 | 1/2 | 3/2 | 2 | ... |
| $\Delta_k$ | 0.5 | 0.6 | 0.7 | ... | ... |
| $\tau_k$ | 0 | 0.25 | 0.85 | 1.20 | ... |
| $y_k$ | 0 | 0.23 | 0.88 | 1.24 | ... |

Table 1: A possible realization from the model: a ritardando. For clarity we assume $\zeta_\tau = 0$.

In the following sections, we will sometimes refer use $z_k = (\tau_k, \Delta_k)^T$ and refer to $z_{0:K}$ as a *tempo trajectory*. Given this definition, we can compactly represent Eq. 4 and Eq. 5 by

$$z_k = \begin{pmatrix} 1 & \gamma_k \\ 0 & 1 \end{pmatrix} z_{k-1} + \zeta_k \tag{7}$$

where $\zeta_k = (\zeta_{\tau_k}, \zeta_{\Delta_k})$.

### 2.3 Extensions

There are several possible extensions to this basic parameterization. For example, one could represent the period $\Delta$ in the logarithmic scale. This warping ensures positivity and seems to be perceptually more plausible since it promotes equal *relative* changes in tempo rather than on an absolute scale (Grubb, 1998; Cemgil et al., 2001). Although the resulting model becomes nonlinear, it can be approximated fairly well by an extended Kalman filter (Bar-Shalom & Li, 1993).

A simple random walk model for tempo fluctuations such as in Eq. 7 seems not to be very realistic. We would expect the tempo deviations to be more structured and smoother.

In our dynamical system framework such smooth deviations can be modeled by increasing the dimensionality of $z$ to include higher order "inertia" variables (Cemgil et al., 2001). For example consider the following model,

$$\begin{pmatrix} \tau_k \\ \Delta_{1,k} \\ \Delta_{2,k} \\ \vdots \\ \Delta_{D-1,k} \end{pmatrix} = \begin{pmatrix} 1 & \gamma_k & \gamma_k & 0 & \dots & 0 \\ 0 & 1 & 0 & 0 & \dots & 0 \\ 0 & 0 & & & & \\ \vdots & \vdots & & A & & \\ 0 & 0 & & & & \end{pmatrix} \begin{pmatrix} \tau_{k-1} \\ \Delta_{1,k-1} \\ \Delta_{2,k-1} \\ \vdots \\ \Delta_{D-1,k-1} \end{pmatrix} + \zeta_k \tag{8}$$

We choose this particular parameterization because we wish to interpret $\Delta_1$ as the slowly varying "average" tempo and $\Delta_2$ as a temporary change in the tempo. Such a model is useful for situations where the performer fluctuates around an almost constant tempo; a random walk model is not sufficient in this case because it forgets the initial values. Additional state variables $\Delta_3, \dots, \Delta_{D-1}$ act like additional "memory" elements. By choosing the parameter matrix $A$ and noise covariance matrix $Q$, one can model a rich range of temporal structures in expressive timing deviations.

The score prior can be improved by using a richer model. For example to allow for different time signatures and alternative rhythmic subdivisions, one can introduce additional hidden variables (See Cemgil et al. (2000) or Appendix A) or use a Markov chain (Raphael, 2001a). Potentially, such extensions make it easier to capture additional structure in musical rhythm (such as "weak" positions are followed more likely by "strong" positions). On the other hand, the number of model parameters rapidly increases and one has to be more cautious in order to avoid overfitting.

For score typesetting, we need to quantize note durations as well, i.e., associate note offsets with quantization locations. A simple way of accomplishing this is to define an indicator sequence $u_{0:K}$ that identifies whether $y_k$ is an onset ($u_k = 1$) or an offset ($u_k = 0$). Given $u_k$, we can redefine the observation model as $p(y_k|\tau_k, u_k) = u_k \mathcal{N}(0, R) + (1 - u_k)\mathcal{N}(0, R_{\text{off}})$ where $R_{\text{off}}$ is the observation noise associated with offsets. A typical model would have $R_{\text{off}} \gg R$. For $R_{\text{off}} \to \infty$, the offsets would have no effect on the tempo process. Moreover, since $u_k$ are always observed, this extension requires just a simple lookup.

In principle, one must allow for arbitrary long intervals between onsets, hence $\gamma_k$ are drawn from an infinite (but discrete) set. In our subsequent derivations, we assume that the number of possible intervals is fixed a-priori. Given an estimate of $z_{k-1}$ and observation $y_k$, almost all of the virtually infinite number of choices for $\gamma_k$ will have almost zero probability and it is easy to identify candidates that would have significant probability mass.

Conceptually, all of the above listed extensions are easy to incorporate into the model and none of them introduces a fundamental computational difficulty to the basic problems of quantization and tempo tracking.

### 2.4 Problem Definition

Given the model, we define rhythm transcription, i.e., quantization as a MAP state estimation problem

$$\gamma^*_{1:K} = \underset{\gamma_{1:K}}{\text{argmax}}\, p(\gamma_{1:K}|y_{0:K}) \tag{9}$$
$$p(\gamma_{1:K}|y_{0:K}) = \int dz_{0:K} p(\gamma_{1:K}, z_{0:K}|y_{0:K})$$

and tempo tracking as a filtering problem

$$z^*_k = \underset{z_k}{\text{argmax}} \sum_{\gamma_{1:k}} p(\gamma_{1:k}, z_k|y_{0:k}) \tag{10}$$

The quantization problem is a smoothing problem: we wish to find the most likely score $\gamma^*_{1:K}$ given all the onsets in the performance. This is useful in "offline" applications such as score typesetting.

For real-time interaction, we need to have an online estimate of the tempo/beat $z_k$. This information is carried forth by the filtering density $p(\gamma_{1:k}, z_k|y_{0:k})$ in Eq.10. Our definition of the best tempo $z^*_k$ as the maximum is somewhat arbitrary. Depending upon the requirements of an application, one can make use of other features of the filtering density. For example, the variance of $\sum_{\gamma_{1:k}} p(\gamma_{1:k}, z_k|y_{0:k})$ can be used to estimate "amount of confidence" in tempo interpretation or $\arg\max_{z_k,\gamma_{1:k}} p(\gamma_{1:k}, z_k|y_{0:k})$ to estimate most likely score-tempo pair so far.

Unfortunately, the quantities in Eq. 9 and Eq. 10 are intractable due to the explosion in the number of mixture components required to represent the exact posterior at each step $k$ (See Figure 2). For example, to calculate the exact posterior in Eq. 9 we need to evaluate the following expression:

$$p(\gamma_{1:K}|y_{0:K}) = \frac{1}{Z}\int dz_{0:K} p(y_{0:K}|z_{0:K}, \gamma_{1:K}) p(z_{0:K}|\gamma_{1:K}) p(\gamma_{1:K}) \tag{11}$$
$$= \frac{1}{Z} p(y_{0:K}|\gamma_{1:K}) p(\gamma_{1:K}) \tag{12}$$

where the normalization constant is given by $Z = p(y_{0:K}) = \sum_{\gamma_{1:K}} p(y_{0:K}|\gamma_{1:K})p(\gamma_{1:K})$. For each trajectory $\gamma_{1:K}$, the integral over $z_{0:K}$ can be computed stepwise in $k$ by the Kalman filter (See appendix B.1). However, to find the MAP state of Eq. 11, we need to evaluate $p(y_{0:K}|\gamma_{1:K})$ independently for each of the exponentially many trajectories. Consequently, the quantization problem in Eq. 9 can only be solved approximately.

For accurate approximation, we wish to exploit any inherent independence structure of the exact posterior. Unfortunately, since $z$ and $c$ are integrated over, all $\gamma_k$ become coupled and in general $p(\gamma_{1:K}|y_{0:K})$ does not possess any conditional independence structure (e.g., a Markov chain) that would facilitate efficient calculation. Consequently, we will resort to numerical approximation techniques.

## 3. Monte Carlo Simulation

Consider a high dimensional probability distribution

$$p(\mathbf{x}) = \frac{1}{Z} p^*(\mathbf{x}) \tag{13}$$

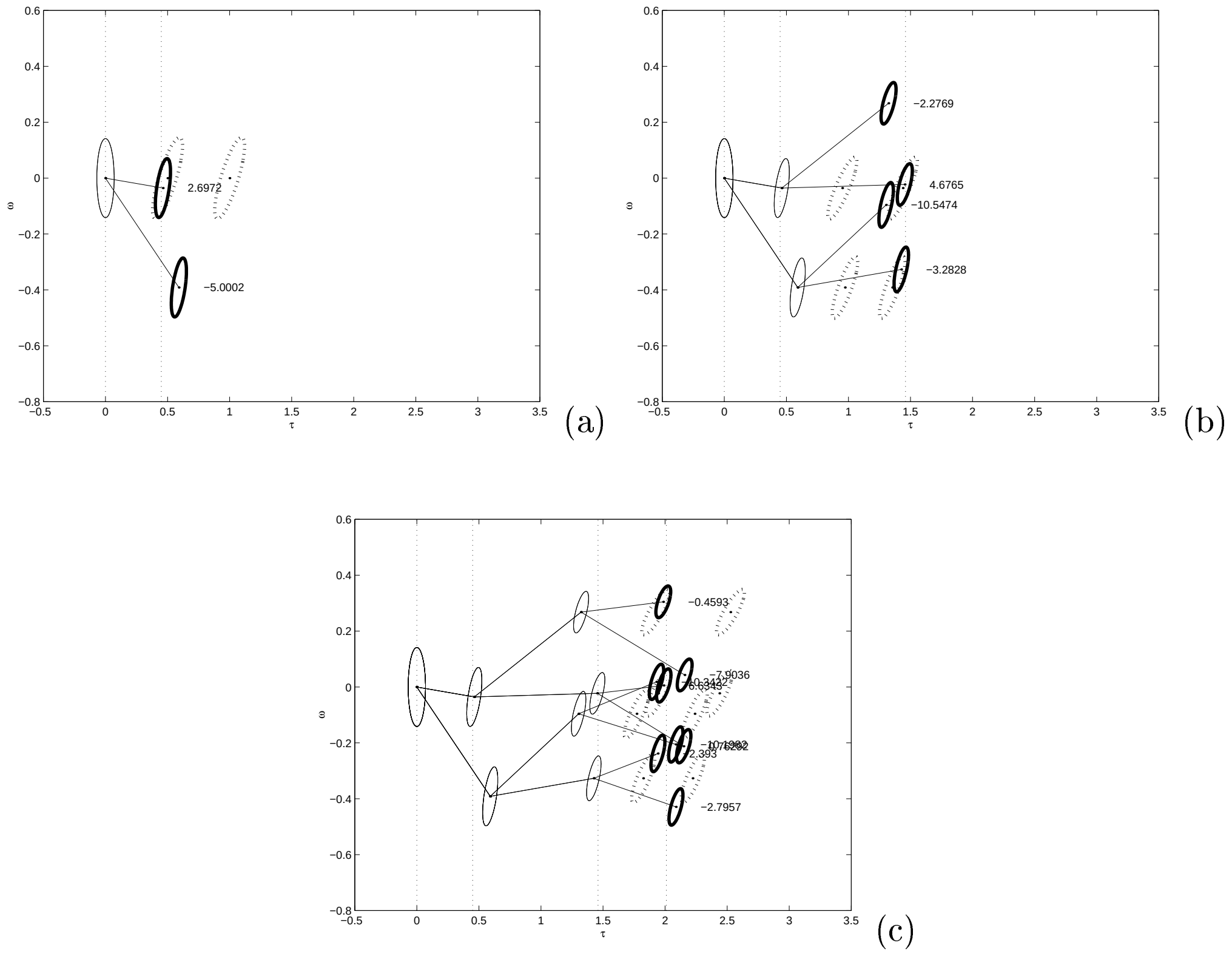

Figure 2: Example demonstrating the explosion of the number of components to represent the exact posterior. Ellipses denote the conditional marginals $p(\tau_k, \omega_k | c_{0:k}, y_{0:k})$. (We show the period in logarithmic scale where $\omega_k = \log_2 \Delta_k$). In this toy example, we assume that a score consists only of notes of length ♪ and ♩, i.e., $\gamma_k$ can be either 1/2 or 1. **(a)** We start with a unimodal posterior $p(\tau_0, \omega_0 | c_0, y_0)$, e.g., a Gaussian centered at $(\tau, \omega) = (0, 0)$. Since we assume that a score can only consist of eight- and quarter notes, i.e., $\gamma_k \in \{1/2,\ 1\}$. the predictive distribution $p(\tau_1, \omega_1 | c_{0:1}, y_0)$ is bimodal where the modes are centered at $(0.5, 0)$ and $(1, 0)$ respectively (shown with a dashed contour line). Once the next observation $y_1$ is observed (shown with a dashed vertical line around $\tau = 0.5$), the predictive distribution is updated to yield $p(\tau_1, \omega_1 | c_{0:1}, y_{0:1})$. The numbers denote the respective log-posterior weight of each mixture component. **(b)** The predictive distribution $p(\tau_2, \omega_2 | c_{0:1}, y_{0:1})$ at step $k = 2$ has now 4 modes, two for each component of $p(\tau_1, \omega_1 | c_{0:1}, y_{0:1})$. **(c)** The number of components grows exponentially with $k$.

where the normalization constant $Z = \int d\mathbf{x} p^*(\mathbf{x})$ is not known but $p^*(\mathbf{x})$ can be evaluated at any particular $\mathbf{x}$. Suppose we want to estimate the expectation of a function $f(\mathbf{x})$ under the distribution $p(\mathbf{x})$ denoted as

$$\langle f(\mathbf{x}) \rangle_{p(\mathbf{x})} = \int dx f(\mathbf{x}) p(\mathbf{x})$$

e.g., the mean of $\mathbf{x}$ under $p(\mathbf{x})$ is given by $\langle \mathbf{x} \rangle$. The intractable integration can be approximated by an average if we can find $N$ points $\mathbf{x}^{(i)}$, $i = 1 \dots N$ from $p(\mathbf{x})$

$$\langle f(\mathbf{x}) \rangle_{p(\mathbf{x})} \approx \frac{1}{N} \sum_{i=1}^{N} f(\mathbf{x}^{(i)}) \tag{14}$$

When $\mathbf{x}^{(i)}$ are generated by independently sampling from $p(\mathbf{x})$, it can be shown that as $N$ approaches infinity, the approximation becomes exact.

However, generating independent samples from $p(\mathbf{x})$ is a difficult task in high dimensions but it is usually easier to generate *dependent* samples, that is we generate $\mathbf{x}^{(i+1)}$ by making use of $\mathbf{x}^{(i)}$. It is somewhat surprising, that even if $\mathbf{x}^{(i)}$ and $\mathbf{x}^{(i+1)}$ are correlated (and provided ergodicity conditions are satisfied), Eq. 14 remains still valid and estimated quantities converge to their true values when number of samples $N$ goes to infinity.

A sequence of dependent samples $\mathbf{x}^{(i)}$ is generated by using a Markov chain that has the stationary distribution $p(\mathbf{x})$. The chain is defined by a collection of transition probabilities, i.e., a *transition kernel* $T(\mathbf{x}^{(i+1)}|\mathbf{x}^{(i)})$. The definition of the kernel is implicit, in the sense that one defines a procedure to generate the $\mathbf{x}^{(i+1)}$ given $\mathbf{x}^{(i)}$. The *Metropolis* algorithm (Metropolis & Ulam, 1949; Metropolis, Rosenbluth, Rosenbluth, Teller, & Teller, 1953) provides a simple way of defining an ergodic kernel that has the desired stationary distribution $p(\mathbf{x})$. Suppose we have a sample $\mathbf{x}^{(i)}$. A candidate $\mathbf{x}'$ is generated by sampling from a symmetric proposal distribution $q(\mathbf{x}'|\mathbf{x}^{(i)})$ (for example a Gaussian centered at $\mathbf{x}^{(i)}$). The candidate $\mathbf{x}'$ is accepted as the next sample $\mathbf{x}^{(i+1)}$ if $p(\mathbf{x}') > p(\mathbf{x}^{(i)})$. If $\mathbf{x}'$ has a lower probability, it can be still accepted, but only with probability $p(\mathbf{x}')/p(\mathbf{x}^{(i)})$. The algorithm is initialized by generating the first sample $x^{(0)}$ according to an (arbitrary) proposal distribution.

However for a given transition kernel $T$, it is hard to assess the time required to converge to the stationary distribution so in practice one has to run the simulation until a very large number of samples have been obtained, (see e.g., Roberts & Rosenthal, 1998). The choice of the proposal distribution $q$ is also very critical. A poor choice may lead to the rejection of many candidates $\mathbf{x}'$ hence resulting in a very slow convergence to the stationary distribution.

For a large class of probability models, where the full posterior $p(\mathbf{x})$ is intractable, one can still efficiently compute marginals of form $p(x_k|\mathbf{x}_{-k})$, $\mathbf{x}_{-k} = x_1 \dots x_{k-1}, x_{k+1}, \dots x_K$ exactly. In this case one can apply a more specialized Markov chain Monte Carlo (MCMC) algorithm, the *Gibbs sampler* given below.

1. Initialize $x_{1:K}^{(0)}$ by sampling from a proposal $q(x_{1:K})$
2. For $i = 0 \dots N-1$

- For $k = 1, \dots, K$, Sample

$$x_k^{(i+1)} \sim p(x_k | x_{1:k-1}^{(i+1)}, x_{k+1:K}^{(i)}) \tag{15}$$

In contrast to the Metropolis algorithm, where the new candidate is a vector $\mathbf{x}'$, the Gibbs sampler uses the exact marginal $p(x_k|\mathbf{x}_{-k})$ as the proposal distribution. At each step, the sampler updates only one coordinate of the current state $\mathbf{x}$, namely $x_k$, and the new candidate is guaranteed to be accepted.

Note that, in principle we don't need to sample $x_k$ sequentially, i.e., we can choose $k$ randomly provided that each slice is visited equally often in the limit. However, a deterministic scan algorithm where $k = 1, \dots K$, provides important time savings in the type of models that we consider here.

### 3.1 Simulated Annealing and Iterative Improvement

Now we shift our focus from sampling to MAP state estimation. In principle, one can use the samples generated by any sampling algorithm (Metropolis-Hastings or Gibbs) to estimate the MAP state $\mathbf{x}^*$ of $p(\mathbf{x})$ by $\underset{i=1:N}{\operatorname{argmax}}\, p(\mathbf{x}^{(i)})$. However, unless the posterior is very much concentrated around the MAP state, the sampler may not visit $\mathbf{x}^*$ even though the samples $\mathbf{x}^{(i)}$ are obtained from the stationary distribution. In this case, the problem can be simply reformulated to sample not from $p(\mathbf{x})$ but from a distribution that is concentrated at local maxima of $p(\mathbf{x})$. One such class of distributions are given by $p_{\rho_j}(\mathbf{x}) \propto p(\mathbf{x})^{\rho_j}$. A sequence of exponents $\rho_1 < \rho_2 < \dots < \rho_j < \dots$ is called to be a *cooling schedule* or *annealing schedule* owing to the inverse temperature interpretation of $\rho_j$ in statistical mechanics, hence the name *Simulated Annealing* (SA) (Aarts & van Laarhoven, 1985). When $\rho_j \to \infty$ sufficiently slowly in $j$, the cascade of MCMC samplers each with the stationary distribution $p_{\rho_j}(\mathbf{x})$ is guaranteed (in the limit) to converge to the global maximum of $p(\mathbf{x})$. Unfortunately, for this convergence result to hold, the cooling schedule must go very slowly (in fact, logarithmically) to infinity. In practice, faster cooling schedules must be employed.

*Iterative improvement* (II) (Aarts & van Laarhoven, 1985) is a heuristic simulated annealing algorithm with a very fast cooling schedule. In fact, $\rho_j = \infty$ for all $j$. The eventual advantage of this greedy algorithm is that it converges in a few iterations to a local maximum. By restarting many times from different initial configurations $\mathbf{x}$, one hopes to find different local maxima of $p(\mathbf{x})$ and eventually visit the MAP state $\mathbf{x}^*$. In practice, by using the II heuristic one may find better solutions than SA for a limited computation time.

From an implementation point of view, it is trivial to convert MCMC code to SA (or II) code. For example, consider the Gibbs sampler. To implement SA, we need to construct a cascade of Gibbs samplers, each with stationary distribution $p(\mathbf{x})^{\rho_j}$. The exact one time slice marginal of this distribution is $p(x_k|\mathbf{x}_{-k})^{\rho_j}$. So, SA just samples from the actual (temperature=1) marginal $p(x_k|\mathbf{x}_{-k})$ raised to a power $\rho_j$.

### 3.2 The Switching State Space Model and MAP Estimation

To solve the rhythm quantization problem, we need to calculate the MAP state of the posterior in Eq. 11

$$p(\gamma_{1:K}|y_{0:K}) \quad \propto \quad p(\gamma_{1:K}) \int dz_{0:K} p(y_{0:K}|z_{0:K}, \gamma_{1:K}) p(z_{0:K}|\gamma_{1:K}) \tag{16}$$

This is a combinatorial optimization problem: we seek the maximum of a function $p(\gamma_{1:K}|y_{0:K})$ that associates a number with each of the discrete configurations $\gamma_{1:K}$. Since it is not feasible to visit all of the exponentially many configurations to find the maximizing configuration $\gamma^*_{1:K}$, we will resort to stochastic search algorithms such as simulated annealing (SA) and iterative improvement (II). Due to the strong relationship between the Gibbs sampler and SA (or II), we will first review the Gibbs sampler for the switching state space model.

The first important observation is that, conditioned on $\gamma_{1:K}$, the model becomes a linear state space model and the integration on $z_{0:K}$ can be computed analytically using Kalman filtering equations. Consequently, one can sample only $\gamma_{1:K}$ and integrate out $z$. The analytical marginalization, called *Rao-Blackwellization* (Casella & Robert, 1996), improves the efficiency of the sampler (e.g., see Doucet, de Freitas, Murphy, & Russell, 2000a).

Suppose now that each switch variable $\gamma_k$ can have $S$ distinct states and we wish to generate $N$ samples (i.e trajectories) $\{\gamma^{(i)}_{1:K}, i = 1 \dots N\}$. A naive implementation of the Gibbs sampler requires that at each step $k$ we run the Kalman filter $S$ times on the whole observation sequence $y_{0:K}$ to compute the proposal $p(\gamma_k|\gamma^{(i)}_{1:k-1}, \gamma^{(i-1)}_{k+1:K}, y_{0:K})$. This would result in an algorithm of time complexity $O(NK^2S)$ that is prohibitively slow when $K$ is large. Carter and Kohn (1996) have proposed a much more time efficient deterministic scan Gibbs sampler that circumvents the need to run the Kalman filtering equations at each step $k$ on the whole observation sequence $y_{0:K}$. See also (Doucet & Andrieu, 2001; Murphy, 2002).

The method is based on the observation that the proposal distribution $p(\gamma_k|\cdot)$ can be factorized as a product of terms that either depend on past observations $y_{0:k}$ or the future observations $y_{k+1:K}$. So the contribution of the future can be computed a-priori by a backward filtering pass. Subsequently, the proposal is computed and samples $\gamma^{(i)}_k$ are generated during the forward pass. The sampling distribution is given by

$$p(\gamma_k|\boldsymbol{\gamma}_{-k}, y_{0:K}) \propto p(\gamma_k|\boldsymbol{\gamma}_{-k})p(y_{0:K}|\gamma_{1:K}) \tag{17}$$

where the first term is proportional to the joint prior $p(\gamma_k|\boldsymbol{\gamma}_{-k}) \propto p(\gamma_k, \boldsymbol{\gamma}_{-k})$. The second term can be decomposed as

$$\begin{aligned} p(y_{0:K}|\gamma_{1:K}) &= \int dz_k p(y_{k+1:K}|y_{0:k}, z_k, \gamma_{1:K})p(y_{0:k}, z_k|\gamma_{1:K}) && (18) \\ &= \int dz_k p(y_{k+1:K}|z_k, \gamma_{k+1:K})p(y_{0:k}, z_k|\gamma_{1:k}) && (19) \end{aligned}$$

Both terms are (unnormalized) Gaussian potentials hence the integral can be evaluated analytically. The term $p(y_{k+1:K}|z_k, \gamma_{k+1:K})$ is an unnormalized Gaussian potential in $z_k$ and can be computed by backwards filtering. The second term is just the filtering distribution $p(z_k|y_{0:k}, \gamma_{1:k})$ scaled by the likelihood $p(y_{0:k}|\gamma_{1:k})$ and can be computed during forward filtering. The outline of the algorithm is given below, see the appendix B.1 for details.

1. Initialize $\gamma^{(0)}_{1:K}$ by sampling from a proposal $q(\gamma_{1:K})$
2. For $i = 1 \dots N$
   - For $k = K - 1, \dots, 0,$

- Compute $p(y_{k+1:K}|z_k, \gamma^{(i-1)}_{k+1:K})$
- For $k = 1, \dots, K$,
  - For $s = 1 \dots S$
    * Compute the proposal

$$p(\gamma_k = s|\cdot) \propto p(\gamma_k = s, \boldsymbol{\gamma}_{-k}) \int dz_k p(y_{0:k}, z_k|\gamma^{(i)}_{1:k-1}, \gamma_k = s) p(y_{k+1:K}|z_k, \gamma^{(i-1)}_{k+1:K})$$

  - Sample $\gamma^{(i)}_k$ from $p(\gamma_k|\cdot)$

The resulting algorithm has a time complexity of $O(NKS)$, an important saving in terms of time. However, the space complexity increases from $O(1)$ to $O(K)$ since expectations computed during the backward pass need to be stored.

At each step, the Gibbs sampler generates a sample from a single time slice $k$. In certain types of "sticky" models, such as when the dependence between $\gamma_k$ and $\gamma_{k+1}$ is strong, the sampler may get stuck in one configuration, moving very rarely. This is due to the fact that most singleton flips end up in low probability configurations due to the strong dependence between adjacent time slices. As an example, consider the quantization model and two configurations $[\dots \gamma_k, \gamma_{k+1} \dots] = [\dots 1, 1 \dots]$ and $[\dots 3/2, 1/2 \dots]$. By updating only a single slice, it may be difficult to move between these two configurations. Consider an intermediate configuration $[\dots 3/2, 1 \dots]$. Since the duration $(\gamma_k + \gamma_{k+1})$ increases, all future quantization locations $c_{k:K}$ are shifted by $1/2$. That may correspond to a score that is heavily penalized by the prior, thus "blocking" the path.

To allow the sampler move more freely, i.e., to allow for more global jumps, one can sample from $L$ slices jointly. In this case the proposal distribution takes the form

$$\begin{aligned} p(\gamma_{k:k+L-1}|\cdot) \quad \propto \quad & p(\gamma_{k:k+L-1}, \boldsymbol{\gamma}_{-(k:k+L-1)}) \times \\ & \int dz_{k+L-1} p(y_{0:k+L-1}, z_{k+L-1}|\gamma^{(i)}_{1:k-1}, \gamma_{k:k+L-1}) p(y_{k+L:K}|z_{k+L-1}, \gamma^{(i-1)}_{k+L:K}) \end{aligned}$$

Similar to the one slice case, terms under the integral are unnormalized Gaussian potentials (on $z_{k+L-1}$) representing the contribution of past and future observations. Since $\gamma_{k:k+L-1}$ has $S^L$ states, the resulting time complexity for generating $N$ samples is $O(NKS^L)$, thus in practice $L$ must be kept rather small. One remedy would be to use a Metropolis-Hastings algorithm with a heuristic proposal distribution $q(\gamma_{k:k+L-1}|y_{0:K})$ to circumvent exact calculation, but it is not obvious how to construct such a $q$.

One other shortcoming of the Gibbs sampler (and related MCMC methods) is that the algorithm in its standard form is inherently offline; we need to have access to all of the observations $y_{0:K}$ to start the simulation. For certain applications, e.g., automatic score typesetting, a batch algorithm might be still feasible. However in scenarios that require real-time interaction, such as in interactive music performance or tempo tracking, online methods must be used.

### 3.3 Sequential Monte Carlo

Sequential Monte Carlo, a.k.a. particle filtering, is a powerful alternative to MCMC for generating samples from a target posterior distribution. SMC is especially suitable for application in dynamical systems, where observations arrive sequentially.

The basic idea in SMC is to represent the posterior $p(x_{0:k-1}|y_{0:k-1})$ at time $k-1$ by a (possibly weighted) set of samples $\{x_{0:k-1}^{(i)}, i = 1 \dots N\}$ and extend this representation to $\{(x_{0:k-1}^{(i)}, x_k^{(i)}), i = 1 \dots N\}$ when the observation $y_k$ becomes available at time $k$. The common practice is to use importance sampling.

### 3.3.1 Importance Sampling

Consider again a high dimensional probability distribution $p(\mathbf{x}) = p^*(\mathbf{x})/Z$ with an unknown normalization constant. Suppose we are given a *proposal* distribution $q(\mathbf{x})$ that is close to $p(\mathbf{x})$ such that high probability regions of both distributions fairly overlap. We generate independent samples, i.e., *particles*, $\mathbf{x}^{(i)}$ from the proposal such that $q(\mathbf{x}) \approx \sum_{i=1}^N \delta(\mathbf{x} - \mathbf{x}^{(i)})/N$. Then we can approximate

$$
\begin{aligned}
p(\mathbf{x}) &= \frac{1}{Z}\frac{p^*(\mathbf{x})}{q(\mathbf{x})}q(\mathbf{x}) && (20)\\
&\approx \frac{1}{Z}\frac{p^*(\mathbf{x})}{q(\mathbf{x})}\frac{1}{N}\sum_{i=1}^N \delta(\mathbf{x}-\mathbf{x}^{(i)}) && (21)\\
&\approx \sum_{i=1}^N \frac{w^{(i)}}{\sum_{j=1}^N w^{(j)}}\delta(\mathbf{x}-\mathbf{x}^{(i)}) && (22)
\end{aligned}
$$

where $w^{(i)} = p^*(\mathbf{x}^{(i)})/q(\mathbf{x}^{(i)})$ are the *importance weights*. One can interpret $w^{(i)}$ as correction factors to compensate for the fact that we have sampled from the "incorrect" distribution $q(\mathbf{x})$. Given the approximation in Eq.22 we can estimate expectations by weighted averages

$$
\langle f(\mathbf{x})\rangle_{p(\mathbf{x})} \approx \sum_{i=1}^N \tilde{w}^{(i)} f(\mathbf{x}^{(i)}) \tag{23}
$$

where $\tilde{w}^{(i)} = w^{(i)}/\sum_{j=1}^N w^{(j)}$ are the *normalized importance weights.*

### 3.3.2 Sequential Importance Sampling

Now we wish to apply importance sampling to the dynamical model

$$
p(x_{0:K}|y_{0:K}) \propto \prod_{k=0}^K p(y_k|x_k)p(x_k|x_{0:k-1}) \tag{24}
$$

where $x = \{z, \gamma\}$. In principle one can naively apply standard importance sampling by using an arbitrary proposal distribution $q(x_{0:K})$. However finding a good proposal distribution can be hard if $K \gg 1$. The key idea in *sequential importance sampling* is the sequential construction of the proposal distribution, possibly using the available observations $y_{0:k}$, i.e.,

$$
q(x_{0:K}|y_{0:K}) = \prod_{k=0}^K q(x_k|x_{0:k-1}, y_{0:k})
$$

Given a sequentially constructed proposal distribution, one can compute the importance weight recursively as

$$w_k^{(i)} = \frac{p^*(x_{0:k}^{(i)}|y_{0:k})}{q(x_{0:k}^{(i)}|y_{0:k})} = \frac{p(y_k|x_k^{(i)})p(x_k^{(i)}|x_{0:k-1}^{(i)}, y_{0:k-1})}{q(x_k^{(i)}|x_{0:k-1}^{(i)} y_{0:k})} \frac{p(y_{0:k-1}|x_{0:k-1}^{(i)})p(x_{0:k-1}^{(i)})}{q(x_{0:k-1}^{(i)}|y_{0:k-1})} \quad (25)$$

$$= \frac{p(y_k|x_k^{(i)})p(x_k^{(i)}|x_{0:k-1}^{(i)}, y_{0:k-1})}{q(x_k^{(i)}|x_{0:k-1}^{(i)} y_{0:k})} w_{k-1}^{(i)} \quad (26)$$

The sequential update schema is potentially more accurate than naive importance sampling since at each step $k$, one can generate a particle from a fairly accurate proposal distribution that takes the current observation $y_k$ into account. A natural choice for the proposal distribution is the filtering distribution given as

$$q(x_k|x_{0:k-1}^{(i)} y_{0:k}) = p(x_k|x_{0:k-1}^{(i)}, y_{0:k}) \quad (27)$$

In this case the weight update rule in Eq. 26 simplifies to

$$w_k^{(i)} = p(y_k|x_{0:k-1}^{(i)}) w_{k-1}^{(i)}$$

In fact, provided that the proposal distribution $q$ is constructed sequentially and past sampled trajectories are not updated, the filtering distribution is the optimal choice in the sense of minimizing the variance of importance weights $w^{(i)}$ (Doucet, Godsill, & Andrieu, 2000b). Note that Eq. 27 is identical to the proposal distribution used in Gibbs sampling at $k = K$ (Eq 15). At $k < K$, the SMC proposal does not take future observations into account; so we introduce discount factors $w_k$ to compensate for sampling from the wrong distribution.

#### 3.3.3 Selection

Unfortunately, the sequential importance sampling may be degenerate, in fact, it can be shown that the variance of $w_k^{(i)}$ increases with $k$. In practice, after a few iterations of the algorithm, only one particle has almost all of the probability mass and most of the computation time is wasted for updating particles with negligible probability.

To avoid the undesired degeneracy problem, several heuristic approaches are proposed in the literature. The basic idea is to duplicate or discard particles according to their normalized importance weights. The selection procedure can be deterministic or stochastic. Deterministic selection is usually greedy; one chooses $N$ particles with the highest importance weights. In the stochastic case, called *resampling*, particles are drawn with a probability proportional to their importance weight $w_k^{(i)}$. Recall that normalized weights $\{\tilde{w}_k^{(i)}, i = 1 \ldots N\}$ can be interpreted as a discrete distribution on particle labels $(i)$.

### 3.4 SMC for the Switching State Space Model

The SIS algorithm can be directly applied to the switching state space model by sampling directly from $x_k = (z_k, \gamma_k)$. However, the particulate approximation can be quite poor if $z$

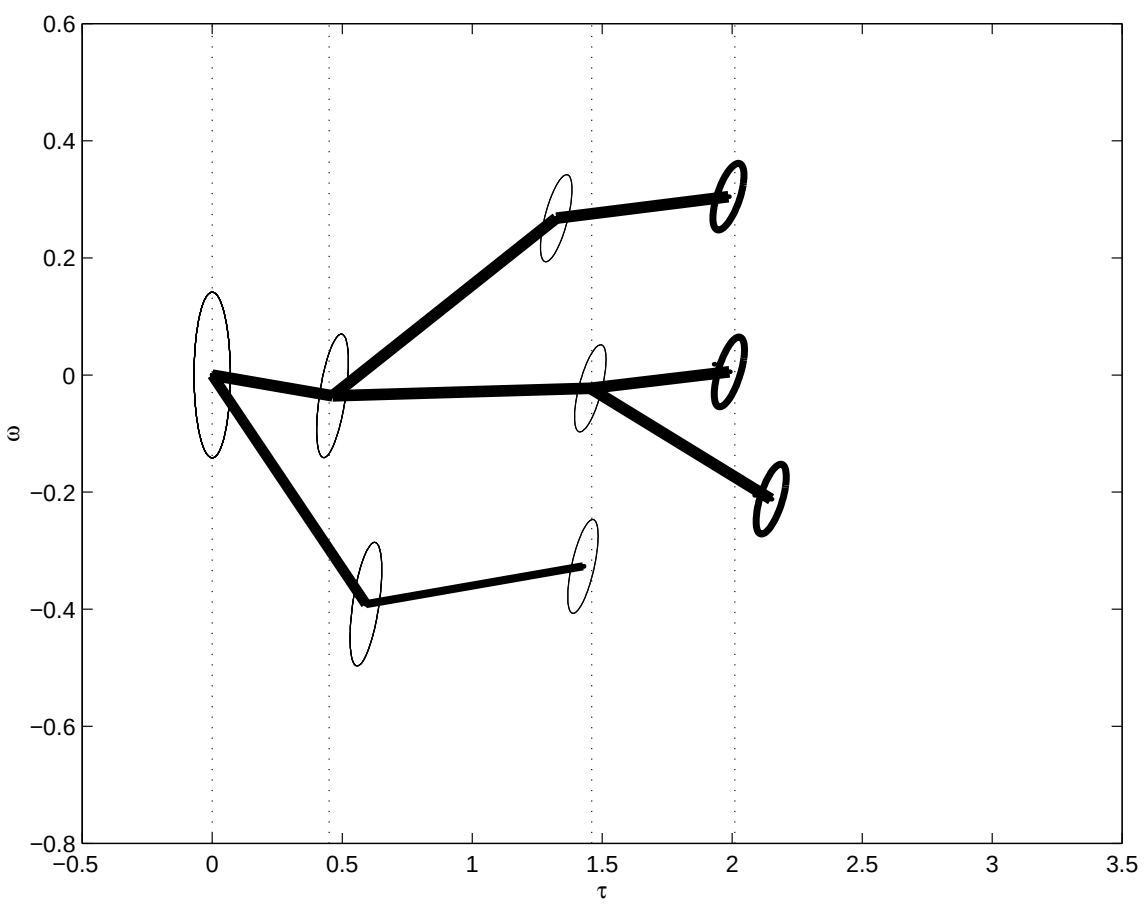

Figure 3: Outline of the algorithm. The ellipses correspond to the conditionals $p(z_k|\gamma_k^{(i)}, y_{0:k})$. Vertical dotted lines denote the observations $y_k$. At each step $k$, particles with low likelihood are discarded. Surviving particles are linked to their parents.

is high dimensional. Hence, too many particles may be needed to accurately represent the posterior.

Similar to the MCMC methods introduced in the previous section, efficiency can be improved by analytically integrating out $z_{0:k}$ and only sampling from $\gamma_{1:k}$. In fact, this form of Rao-Blackwellization is reported to give superior results when compared to standard particle filtering where both $\gamma$ and $z$ are sampled jointly (Chen & Liu, 2000; Doucet et al., 2000b). The improvement is perhaps not surprising, since importance sampling performs best when the sampled space is low dimensional.

The algorithm has an intuitive interpretation in terms of a randomized breadth first tree search procedure: at each new step $k$, we expand $N$ kernels to obtain $S \times N$ new kernels. Consequently, to avoid explosion in the number of branches, we select $N$ out of $S \times N$ branches proportional to the likelihood, See Figure 3. The derivation and technical details of the algorithm are given in the Appendix C.

The tree search interpretation immediately suggests a deterministic version of the algorithm where one selects (without replacement) the $N$ branches with highest weight. We will refer to this method as a *greedy filter* (GF). The method is also known as *split-track* filter (Chen & Liu, 2000) and is closely related to Multiple Hypothesis Tracking (MHT) (Bar-Shalom & Fortmann, 1988). One problem with the greedy selection schema of GF is the loss of particle diversity. Even if the particles are initialized to different locations in $z_0$, (e.g., to different initial tempi), mainly due to the discrete nature of the state space of $\gamma_k$, most of the particles become identical after a few steps $k$. Consequently, results can not be improved by increasing the number of particles $N$. Nevertheless, when only very few particles can be used, say e.g., in a real time application, GF may still be a viable choice.

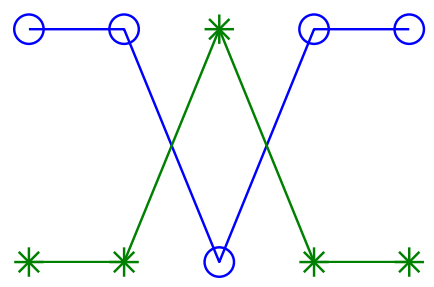

Figure 4: A hypothetical situation where neither of the two particles $\gamma_{1:5}^{(i)}$ is optimal. We would obtain eventually a higher likelihood configuration by interchanging $\gamma_3$ between particles.

### 3.5 SMC and estimation of the MAP trajectory

Like MCMC, SMC is a sampling method. Hence comments made in Section 3.1 about the eventual suboptimality of estimating the MAP trajectory from particles as $\arg\max p(\gamma_{1:K}^{(i)}|y_{0:K})$ also apply here. An hypothetical situation is shown in figure 4.

One obvious solution is to employ the SA "trick" and raise the proposal distribution to a power $p(\gamma_k|\cdot)^\gamma$. However, such a proposal will be peaked on a very few $\gamma$ at each time slice. Consequently, most of the particles will become identical in time and the algorithm eventually degenerates to greedy filtering.

An algorithm for estimating the MAP trajectory from a set of SMC samples is recently proposed in the literature (Godsill, Doucet, & West, 2001). The algorithm relies on the observation that once the particles $x_k^{(i)}$ are sampled during the forward pass, one is left with a discrete distribution defined on the (discrete) support $X_{1:K} = \bigotimes_{k=1}^{K} X_k$. Here $X_k$ denotes is the support of the filtering distribution a time $k$ and $\bigotimes$ is the Cartesian product between sets. Formally, $X_k$ is the set of *distinct* samples at time $k$ and is given by $X_k = \bigcup_i \{x_k^{(i)}\}$.

The distribution $p(X_{1:K}|y_{1:K})^3$ is Markovian because the original state transition model is Markovian, i.e., the posterior can be represented exactly by

$$p(X_{1:K}|y_{1:K}) \propto \prod_{k=1}^{K} p(y_k|X_k)p(X_k|X_{k-1})$$

Consequently, one can find the best MAP trajectory $\arg\max p(X_{1:K})$ by using an algorithm that is analogous to the Viterbi algorithm for hidden Markov models (Rabiner, 1989).

However, this idea does not carry directly to the case when one applies Rao-Blackwellization. In general, when a subset of the hidden variables is integrated out, all time slices of the posterior $p(\Gamma_{1:K}|y_{1:k})$ are coupled, where $\Gamma_{1:K} = \bigotimes_{k=1}^{K} \Gamma_k$ and $\Gamma_k = \bigcup_i \{\gamma_k^{(i)}\}$. One can still employ a chain approximation and run Viterbi, (e.g., Cemgil & Kappen, 2002), but this does not guarantee to find $\arg\max p(\Gamma_{1:K}|y_{1:k})$.

On the other hand, because $\gamma_k^{(i)}$ are drawn from a discrete set, several particles become identical so $\Gamma_k$ has usually a small cardinality when compared to the number of particles $N$. Consequently, it becomes feasible to employ SA or II on the reduced state space $\Gamma_{1:K}$; possibly using a proposal distribution that extends over several time slices $L$.

3. By a slight abuse of notation we use the symbol $X_k$ both as a set and as a general element when used in the argument of a density, $p(y_k|X_k)$ means $p(y_k|x_k)$ s.t. $x_k \in X_k$

In practice, for finding the MAP solution from the particle set $\{\gamma_{1:K}^{(i)}, i = 1 \dots N\}$, we propose to find the best trajectory $i^* = \arg\max_i p(y_{0:K}|\gamma_{1:K}^{(i)})p(\gamma_{1:K}^{(i)})$ and apply iterative improvement starting from the initial configuration $\gamma_{1:K}^{(i^*)}$.

## 4. Simulations

We have compared the inference methods in terms of the quality of the solution and execution time. The tests are carried out both on artificial and real data.

Given the true notation $\gamma_{1:K}^{\text{true}}$, we measure the quality of a solution in terms of the log-likelihood difference

$$\Delta\mathcal{L} = \log \frac{p(y_{0:K}|\gamma_{1:K})p(\gamma_{1:K})}{p(y_{0:K}|\gamma_{1:K}^{\text{true}})p(\gamma_{1:K}^{\text{true}})}$$

and in terms of *edit distance*

$$e(\gamma_{1:K}) = \sum_{k=1}^{K}(1 - \delta(\gamma_k - \gamma_k^{\text{true}}))$$

The edit distance $e(\gamma_{1:K})$ gives simply the number of notes that are quantized wrongly.

### 4.1 Artificial data: Clave pattern

The synthetic example is a repeating "son-clave" pattern 𝄆 𝄽 ♩ ♩ 𝄽 | ♩. ♩. ♩ 𝄇($c = [1, 2, 4, 5.5, 7 \dots]$) with fluctuating tempo. We repeat the pattern 6 times and obtain a score $\gamma_{1:K}$ with $K = 30$.

Such syncopated rhythms are usually hard to transcribe and make it difficult to track the tempo even for experienced human listeners. Moreover, since onsets are absent at prominent beat locations, standard beat tracking algorithms usually loose track.

Given score $\gamma_{1:K}$, we have generated 100 observation sequences $y_{0:K}$ by sampling from the tempo model in Eq. 7. We have parameterized the observation noise variance[4] as $Q = \gamma_k Q_a + Q_b$. In this formulation, the variance depends on the length of the interval between consecutive onsets; longer notes in the score allow for more tempo and timing fluctuation. For the tests on the clave example we have not used a prior model that reflects true source statistics, instead, we have used the generic prior model defined in Section 2.1 with $\lambda = 1$.

All the example cases are sampled from the same score (clave pattern). However, due to the use of the generic prior (that does not capture the exact source statistics well) and a relatively broad noise model, the MAP trajectory $\gamma_{1:K}^*$ given $y_{0:K}$ is not always identical to the original clave pattern. For the $i$'th example, we have defined the "ground truth" $\gamma_{1:K}^{\text{true},i}$ as the highest likelihood solution found using any sampling technique during any independent run. Although this definition of the ground truth introduces some bias, we have found this exercise more realistic as well as more discriminative among various methods when compared to, e.g.,, using a dataset with essentially shorter sequences where the exact MAP

4. The noise covariance parameters were $R = 0.02^2$, $Q_a = 0.06^2 I$ and $Q_b = 0.02^2 I$. I is a $2 \times 2$ identity matrix.

trajectory can be computed by exhaustive enumeration. The wish to stress that the main aim of the simulations on synthetic dataset is to compare effectiveness of different inference techniques; we postpone the actual test whether the model is a good one to our simulations on real data.

We have tested the MCMC methods, namely Gibbs sampling (Gibbs), simulated annealing (SA) and iterative improvement (II) with one and two time slice optimal proposal and for 10 and 50 sweeps. For each onset $y_k$, the optimal proposal $p(\gamma_k|\cdot)$ is computed always on a fixed set, $\Gamma = \{0, 1/4, 2/4 \ldots 3\}$. Figure 6 shows a typical run of MCMC.

Similarly, we have implemented the SMC for $N = \{1, 5, 10, 50, 100\}$ particles. The selection schema was random drawing from the optimal proposal $p(\gamma_k|\cdot)$ computed using one or two time slices. Only in the special case of greedy filtering (GF), i.e., when $N = 1$, we have selected the switch with maximum probability. An example run is shown in Figure 5.

We observe that on average SMC results are superior to MCMC (Figure 7). We observe that, increasing the number of sweeps for MCMC does not improve the solution significantly. On the other hand, increasing the number of particles seems to improve the quality of the SMC solution monotonically. Moreover, the results suggest that sampling from two time slices jointly (with the exception of SA ) does not have a big effect. GF outperforms a particle filter with 5 particles that draws randomly from the proposal. That suggests that for PF with a small number of particles $N$, it may be desirable to use a hybrid selection schema that selects the particle with maximum weight automatically and randomly selects the remaining $N - 1$.

We compare inference methods in terms of execution time and the quality of solutions (as measured by edit distance). As Figure 8 suggests, using a two slice proposal is not justified. Moreover it seems that for comparable computational effort, SMC tends to outperform all MCMC methods.

### 4.2 Real Data: Beatles

We evaluate the performance of the model on polyphonic piano performances. 12 pianists were invited to play two Beatles songs, Michelle and Yesterday. Both pieces have a relatively simple rhythmic structure with ample opportunity to add expressiveness by fluctuating the tempo. The original score is shown in Figure 9(a). The subjects had different musical education and background: four professional jazz players, four professional classical performers and four amateur classical pianists. Each arrangement had to be played in three tempo conditions, three repetitions per tempo condition. The tempo conditions were normal, slow and fast tempo, all in a musically realistic range and all according to the judgment of the performer. Further details are reported in (Cemgil et al., 2001).

#### 4.2.1 Preprocessing

The original performances contained several errors, such as missing notes or additional notes that were not on the original score. Such errors are eliminated by using a matching technique (Heijink, Desain, & Honing, 2000) based on dynamical programming. However, visual inspection of the resulting dataset suggested still several matching errors that we interpret as outliers. To remove these outliers, we have extended the quantization model with a two state switching observation model, i.e., the discrete space consists of $(\gamma_k, i_k)$. In this simple

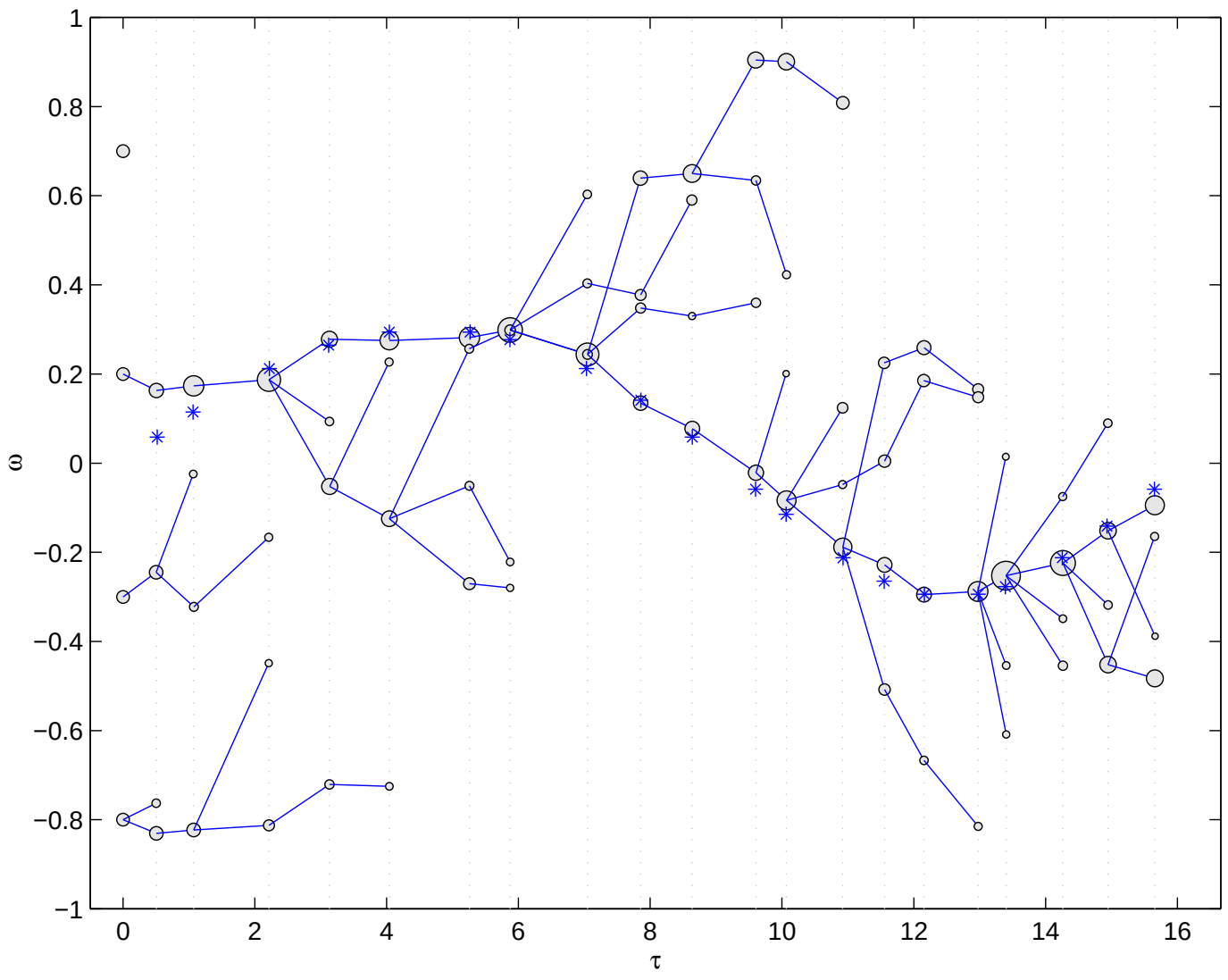

Figure 5: Particle filtering on clave example with 4 particles. Each circle denotes the mean $(\tau_k^{(n)}, \omega_k^{(n)})$ where $\omega_k^{(n)} = \log_2 \Delta_k$. The diameter of each particle is proportional to the normalized importance weight at each generation. '*' denote the true $(\tau, \omega)$ pairs; here we have modulated the tempo deterministically according to $\omega_k = 0.3 \sin(2\pi c_k/32)$, observation noise variance is $R = 0.025^2$.

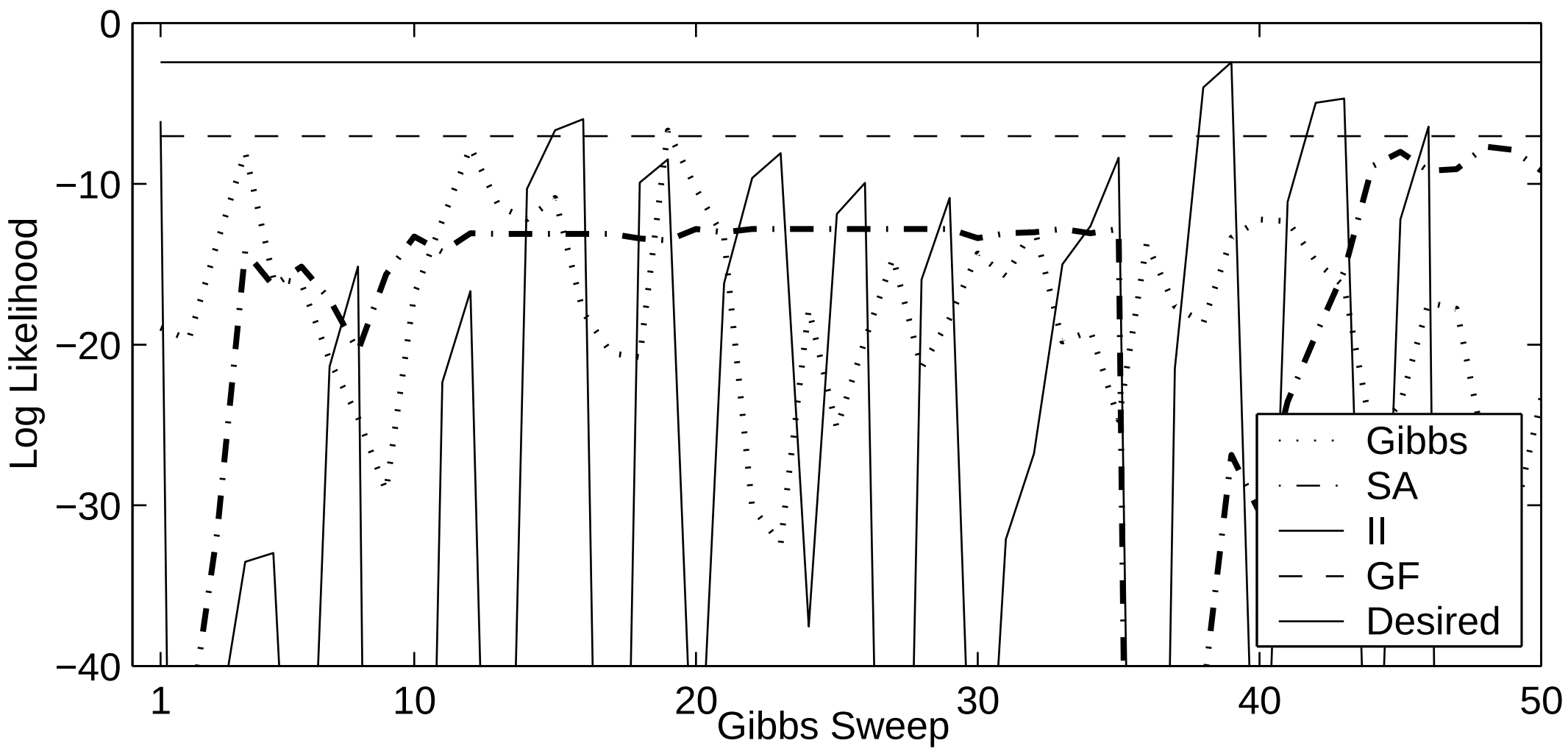

Figure 6: Typical runs of Gibbs sampling, Simulated Annealing (SA) and Iterative Improvement (II) on clave example. All algorithms are initialized to the greedy filter solution. The annealing schedule for SA was linear from $\rho_1 = 0.1$ to $\rho_{33} = 10$ and than proceeding deterministically by $\rho_{34:50} = \infty$. When SA or II converge to a configuration, we reinitialize by a particle filter with one particle that draws randomly proportional to the optimal proposal. Sharp drops in the likelihood correspond to reinitializations. We see that, at the first sweep, the greedy filter solution can only be slightly improved by II. Consequently the sampler reinitializes. The likelihood of SA drops considerably, mainly due to the high temperature, and consequently stabilizes at a suboptimal solution. The Gibbs sampler seems to explore the support of the posterior but is no able to visit the MAP state in this run.

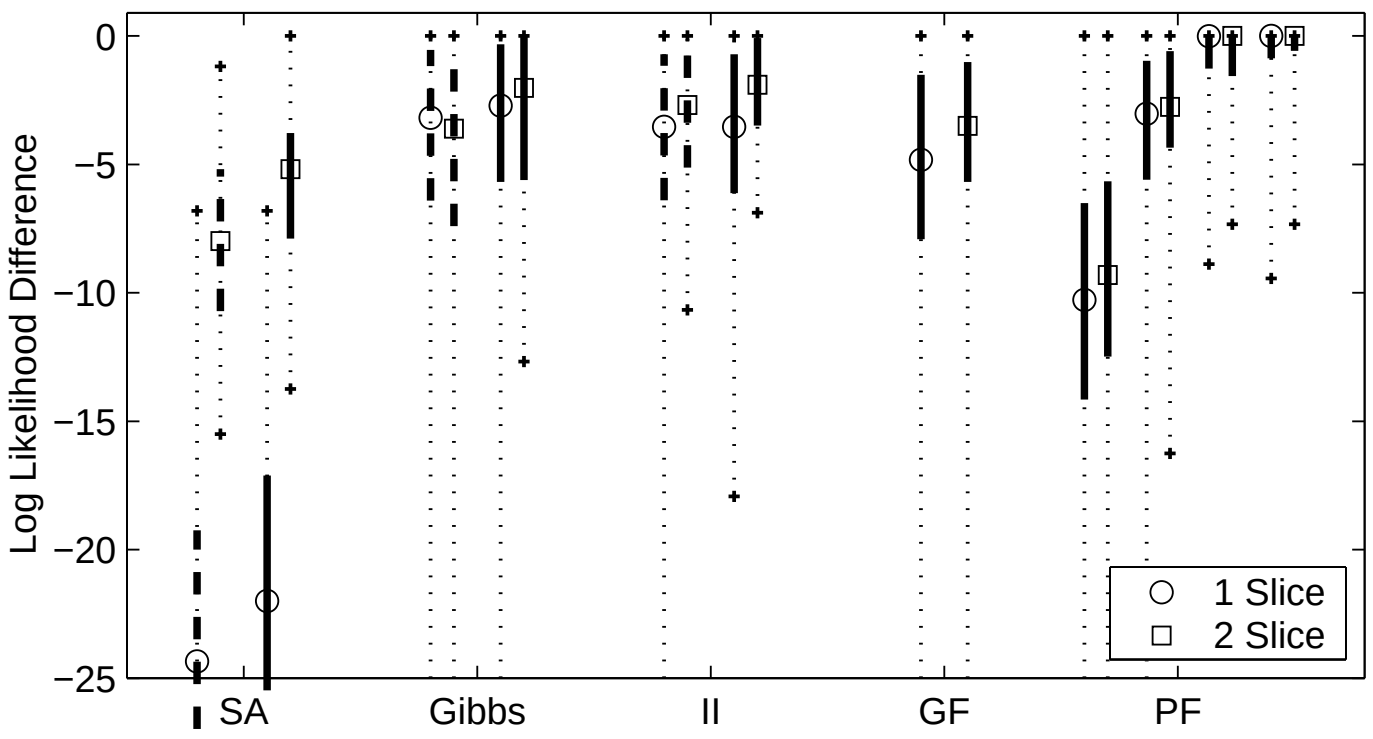

(a) Likelihood Difference

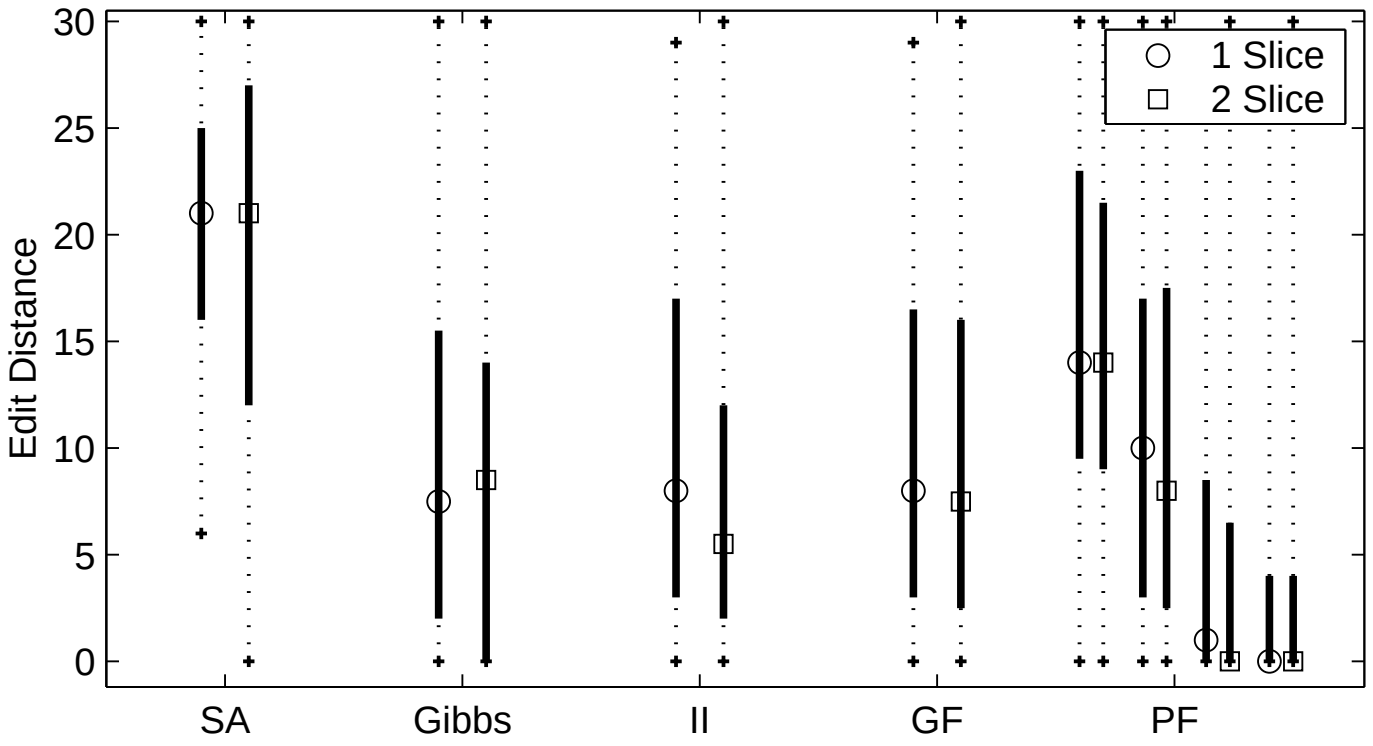

(b) Edit Distance. MCMC results with 10 sweeps are omitted.

Figure 7: Comparison of inference methods on the clave data. The squares and ovals denote the median and the vertical bars correspond to the interval between %25 and %75 quantiles. We have tested the MCMC methods (Gibbs, SA and II) independently for 10 and 50 (shown from left to right). The SMC methods are the greedy filter (GF) and particle filter (PF). We have tested filters with $N = \{5, 10, 50, 100\}$ particles independently (shown from left to right.).

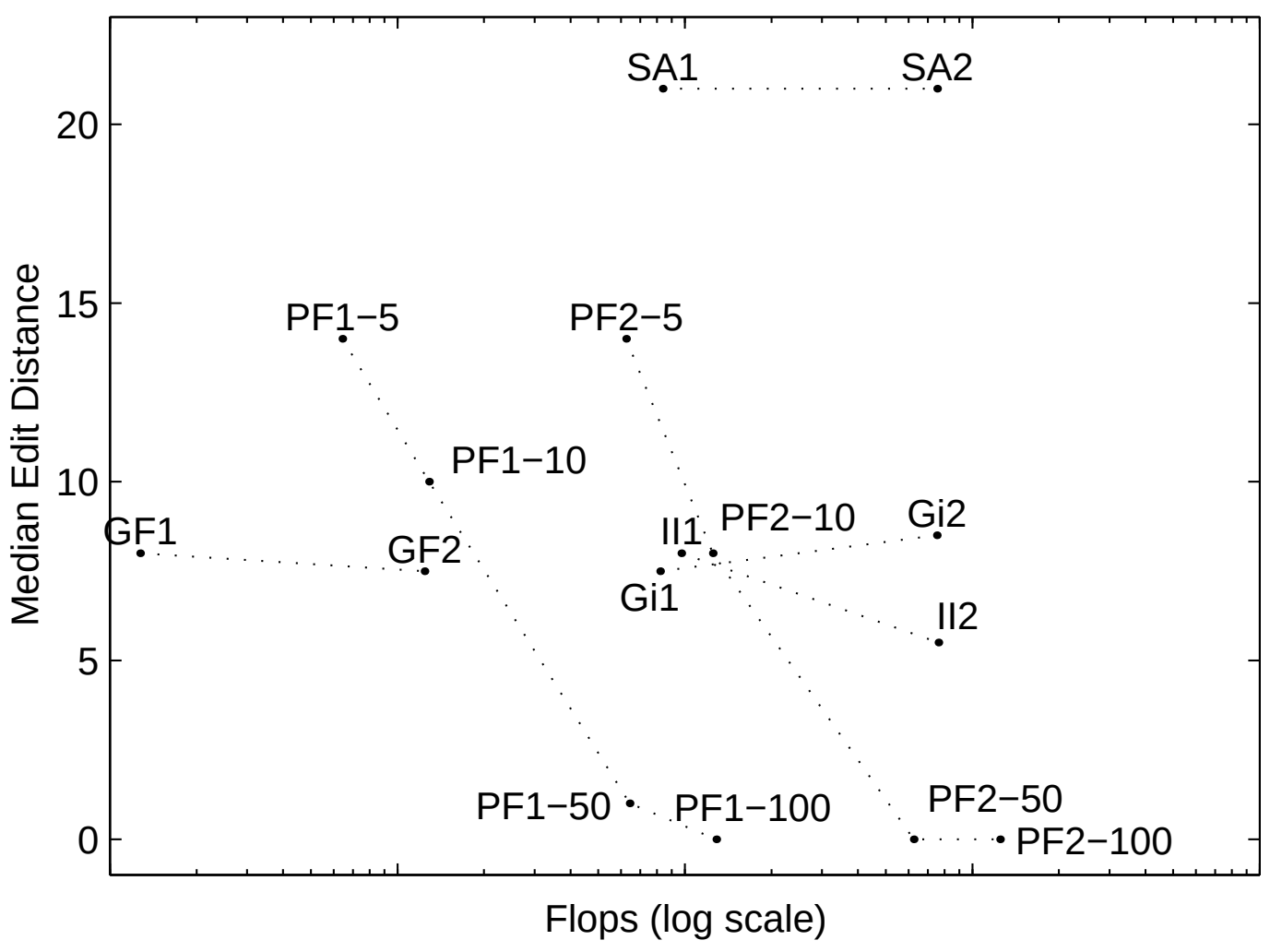

Figure 8: Comparison of execution time in terms of floating point operations. For all methods, the first number (1 or 2) denotes the number slices used by the optimal proposal distribution. For the particle filter (PF), the second number denotes the number of particles. The dashed lines are merely used to connect related methods.

outlier detection mechanism, each switch $i_k$ is a binary indicator variable specifying whether the onset $y_k$ is an outlier or not. We assume that all indicators are independent a-priori and have a uniform prior. The observation model is given by $p(y_k|i_k, \tau_k) = \mathcal{N}(0, R_{i_k})$ [5]. Since the score $\gamma_{1:K}$ is known, the only unknown discrete quantities are the indicators $i_{0:K}$. We have used greedy filtering followed by iterative improvement to find the MAP state of indicators $i_{0:K}$ and eliminated outliers in our further studies. For many performances, there were around $2-4$ outliers, less than 1% of all the notes. The resulting dataset can be downloaded from the url `http://www.snn.kun.nl/~cemgil`.

#### 4.2.2 Parameter Estimation

We have trained tempo tracking models with different dimensionality $D$, where $D$ denotes the dimension of the hidden variable $z$. In all of the models, we use a transition matrix that has the form in Eq. 8.

Since the true score is known, i.e., the quantization location $c_k$ of each onset $y_k$ is given, we can clamp all the discrete variables in the model. Consequently, we can estimate the observation noise variance $R$, the transition noise variance $Q$ and the transition matrix coefficients $A$ from data.

We have optimized the parameters by Expectation-Maximization (EM) for the linear dynamical systems (Shumway & Stoffer, 1982; Ghahramani & Hinton, 1996) using all perfor-

5. We took $R_{i_k=0} = 0.002$ and $R_{i_k=1} = 2$.

mances of "Yesterday" as training data. Similarly, the score prior parameters are estimated by frequency counts from the score of "Yesterday" [6] . All tests are carried out on "Michelle".

### 4.2.3 Results

In Figure 9 we show the result of typesetting a performance with and without tempo tracking. Due to fluctuations in tempo, the quality of the automatically generated score is very poor. The quality can be significantly improved by using our model.

Figure 10 shows some tempo tracking examples on Michelle dataset for pianists from different background and training. We observe that in most cases the results are satisfactory.

In Figure 11, we give a summary of test results on Michelle data in terms of the log-likelihood and edit distance as a function of model order and number of particles used for inference. Figure 11(a) shows that the median likelihood on test data is increasing with model order. This suggests that a higher order filter is able to capture structure in pianists' expressive timing. Moreover, as for the sythetic data, we see a somewhat monotonic increase in the likelihood of solutions found when using more particles.

The edit distance between the original score and the estimates are given in Figure 11(b). Since both pieces are arranged for piano, due to polyphony, there are many onsets that are associated with the same quantization location. Consequently, many $\gamma_k^{\text{true}}$ in the original score are effectively zero. In such cases, typically, the corresponding inter onset interval $y_k - y_{k-1}$ is also very small and the correct quantization (namely $\gamma_k = 0$) can be identified even if the tempo estimate is completely wrong. As a consequence, the edit distance remains small. To make the task slightly more challenging, we exclude the onsets with $\gamma_k^{\text{true}} = 0$ from edit distance calculation.

We observe that the extra prediction ability obtained using a higher order model does not directly translate to a better transcription. The errors are around 5% for all models. On the other hand, the variance of edit distance for higher order models is smaller. This suggests that higher order models tend to be more robust against divergence from the original tempo track.

## 5. Discussion

We have presented a switching state space model for joint rhythm quantization and tempo tracking. The model describes the rhythmic structure of musical pieces by a prior distribution over quantization locations. In this representation, it is easy to construct a generic prior that prefers simpler notations and to learn parameters from a data set. The prior on quantization locations $c_{0:K}$ translates to a non-Markovian distribution over a score $\gamma_{1:K}$.

Timing deviations introduced by performers (tempo fluctuation, accentuations and motor errors) are modeled as independent Gaussian noise sources. Performer specific timing preferences are captured by the parameters of these distributions.

Given the model, we have formulated rhythm quantization as a MAP state estimation problem and tempo tracking as a filtering problem. We have introduced Markov chain

6. The maximum likelihood parameters for a model of dimension $D = 3$ are found to be: $a = -0.072$, $R = 0.013^2$ and $q_\tau = 0.008^2$, $q_{\Delta_1} = 0.007^2$ and $q_{\Delta_2} = 0.050^2$. The prior $p(c)$ is $p(0) = 0.80$, $p(1/3) = 0.0082$, $p(1/2) = 0.15$ $p(5/6) = 0.0418$. Remaining $p(c)$ are set to $10^{-6}$.

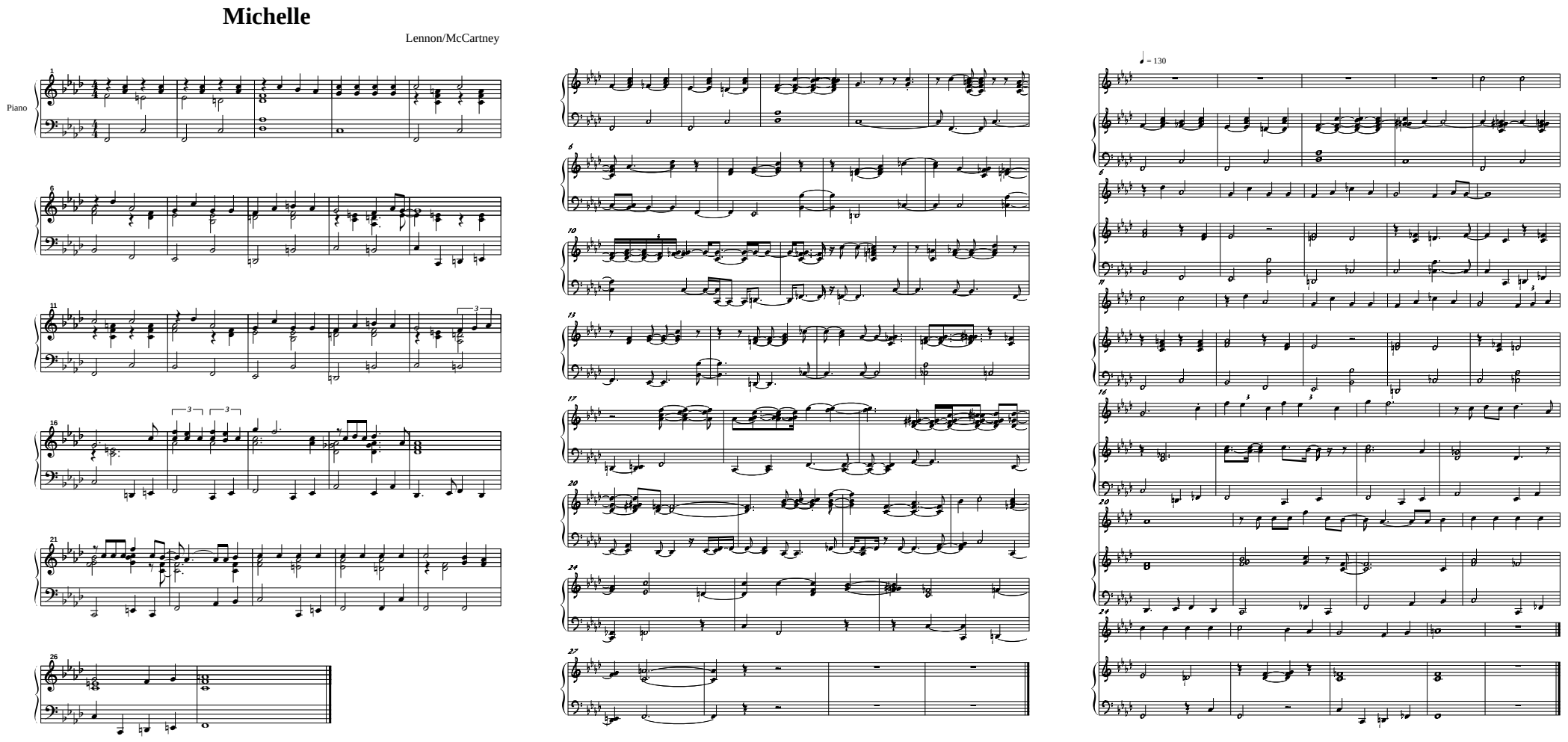

(a) Original Score

(b) Typesetting without processing by the model. Due to fluctuations in tempo, the quality of the score is poor.

(c) Typesetting after tempo tracking and quantization with a particle filter.

Figure 9: Results of Typesetting the scores.

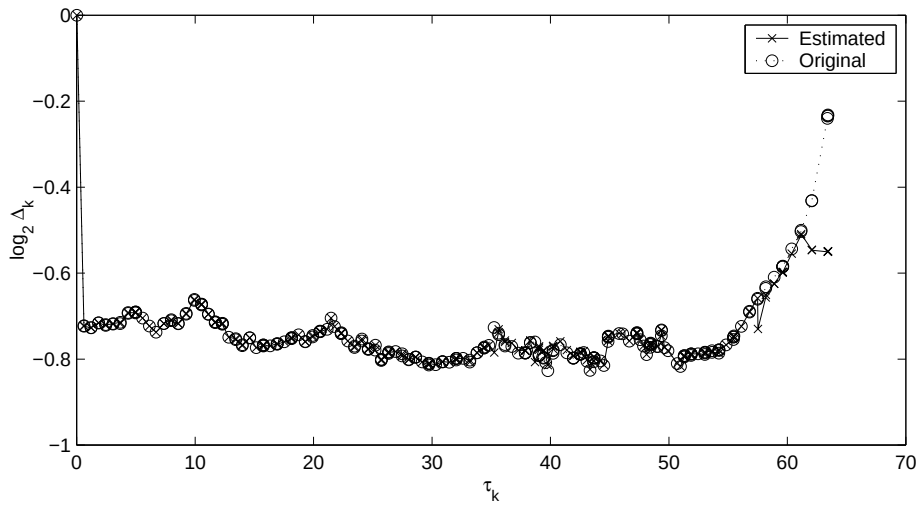

(a) Professional Jazz Pianist

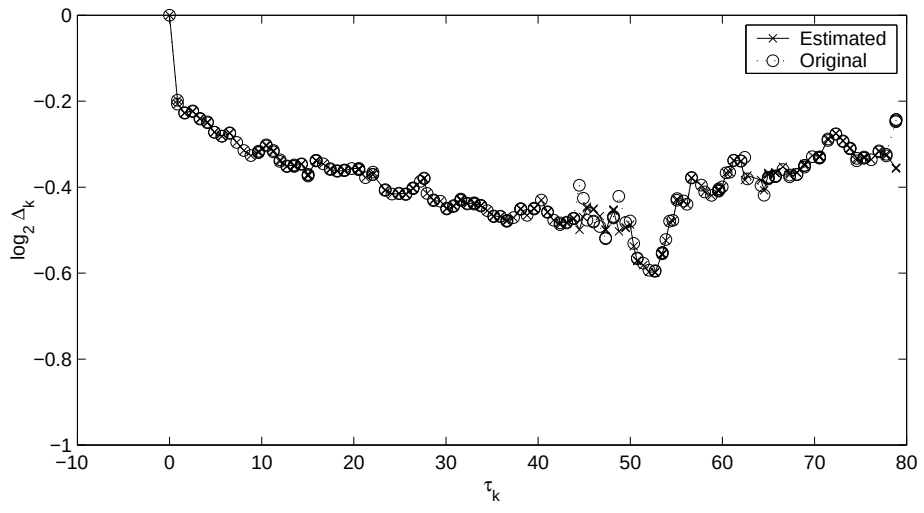

(b) Amateur

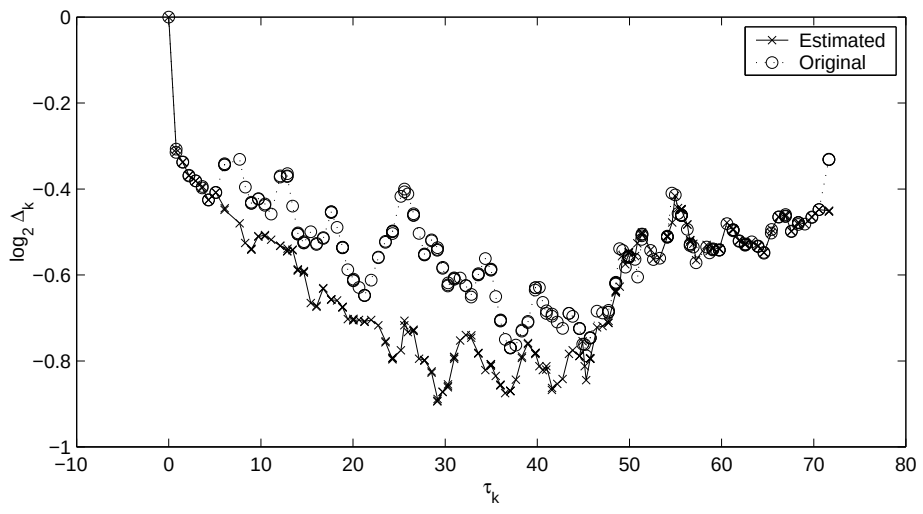

(c) Professional Classical Pianist. The filter temporarily loses track.

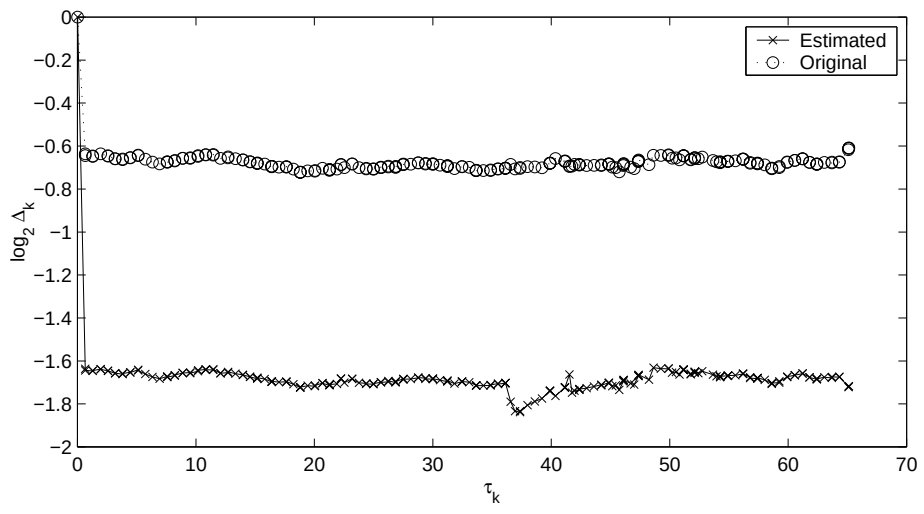

(d) Tracking at twice the rate of the original tempo.

Figure 10: Examples of filtered estimates of $z_{0:K} = [\tau_k, \Delta_k]^T$ from the Beatles data set. Circles denote the mean of $p(z_k|\gamma_{1:k}^{\text{original}}, y_{0:k})$ and "x" denote mean $p(z_k|\gamma_{1:k}^*, y_{0:k})$ obtained by SMC. It is interesting to note different timing characteristics. For example the classical pianist uses a lot more tempo fluctuation than the professional jazz pianist. Jazz pianist slows down dramatically at the end of the piece, the amateur "rushes", i.e., constantly accelerates at the beginning. The tracking and quantization results for (a) and (b) are satisfactory. In (a), the filter loses track at the last two notes, where the pianist dramatically slows down. In (c), the filter loses track but catches up again. In (d), the filter jumps to a metrical level that is twice as fast as the original performance. That would translate to a duplication in note durations only.

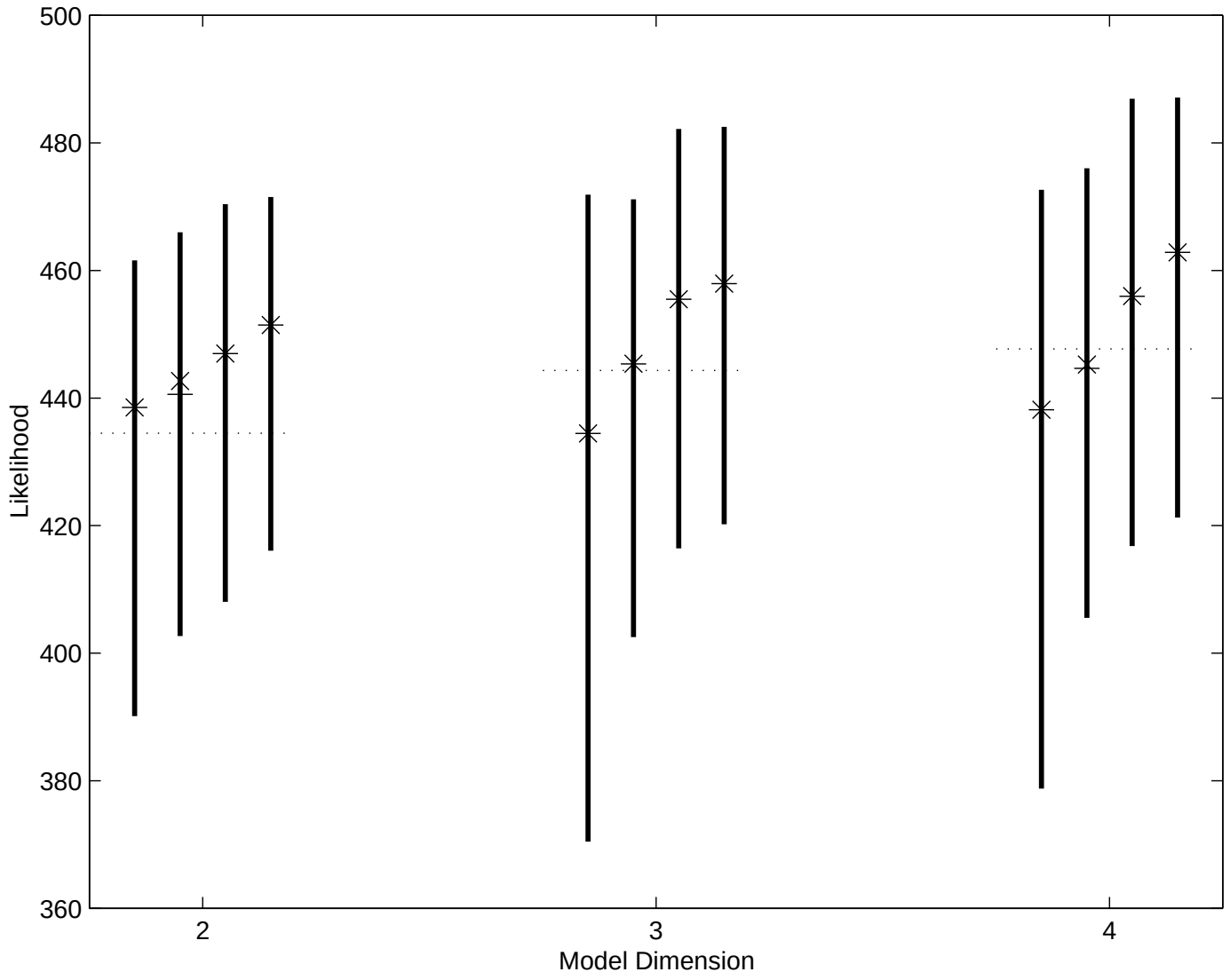

(a) Likelihood. The dashed horizontal line shows the median likelihood of the original score of Michelle under each model.

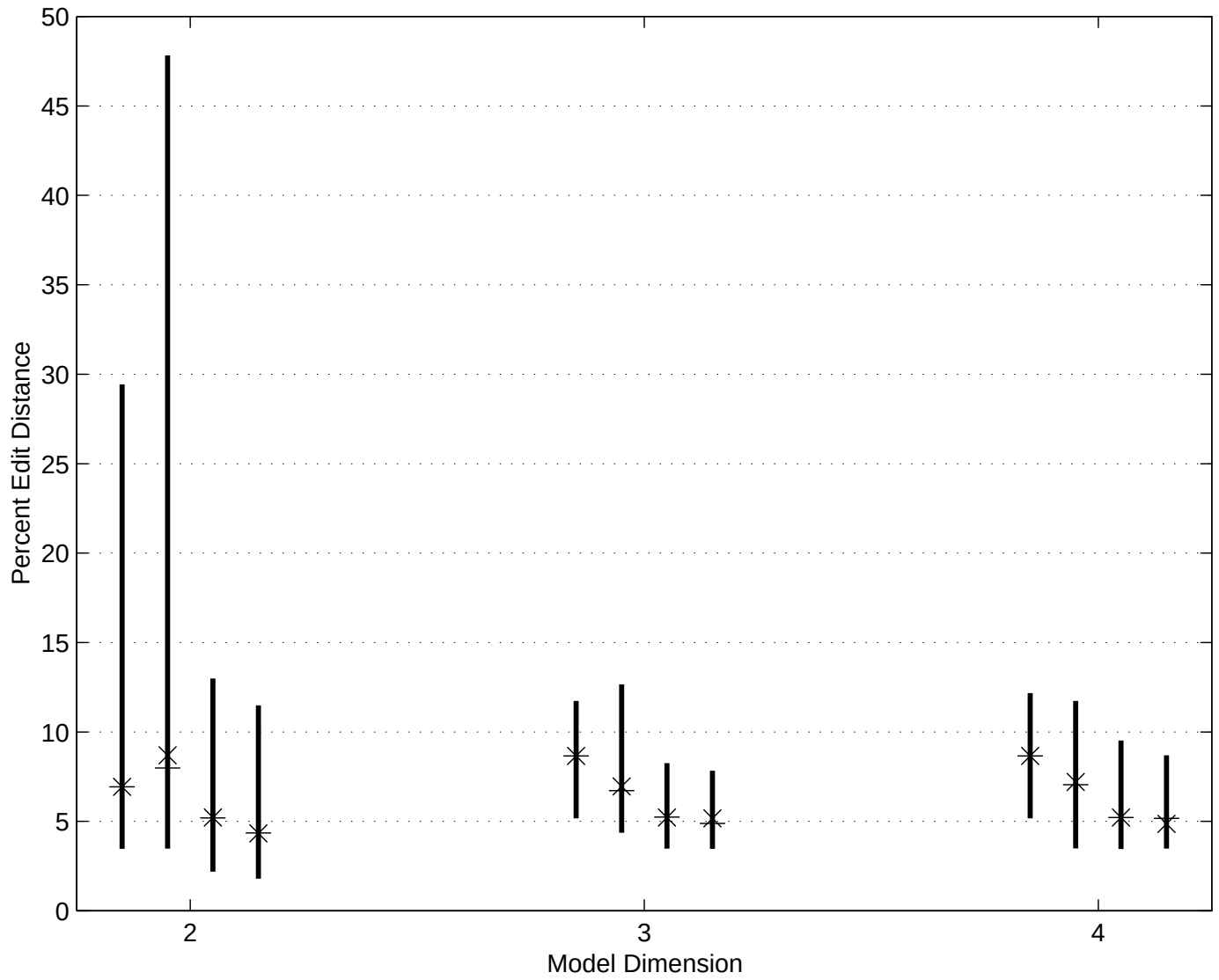

(b) Edit Distance

Figure 11: SMC results on the test data (108 performances of Michelle). For each model we show the results obtained with $N = 1, 10, 20$ and 50 particles. The "-" show the median of the best particle and "x" denote the median after applying iterative improvement. The vertical bars correspond to the interval between %25 and %75 quantiles.

Monte Carlo (MCMC) and sequential Monte Carlo (SMC) to approximate the respective distributions.

The quantization model we propose is similar to that of (Raphael, 2001a). For transcription, Raphael proposes to compute $\arg\max p(c_{0:K}, z_{0:K}|y_{0:K})$ and uses a message propagation scheme that is essentially analogous to Rao-Blackwellized particle filtering. To prevent the number of kernels from explosion, he uses a deterministic selection method, called "thinning". The advantage of Raphael's approach is that the joint MAP trajectory can be computed exactly, provided that the continuous hidden state $z$ is one dimensional and the model is in a parameter regime that keeps the number of propagated Gaussian kernels limited, e.g., if $R$ is small, thinning can not eliminate many kernels. One disadvantage is that the number of kernels varies depending upon the features of the filtering distribution; it is difficult to implement such a scheme in real time. Perhaps more importantly, simple extensions such as increasing the dimensionality of $z$ or introducing nonlinearities to the transition model would render the approach quickly invalid. In contrast, Monte Carlo methods provide a generic inference technique that allow great flexibility in models one can employ.

We have tested our method on a challenging artificial problem (clave example). SMC has outperformed MCMC in terms of the quality of solutions, as measured in terms of the likelihood as well as the edit distance. We propose the use of SMC for both problems. For finding the MAP quantization, we propose to apply iterative improvement (II) to the SMC solution on the reduced configuration space.

The correct choice of the score prior is important in the overall performance of the system. Most music pieces tend to have a certain rhythmical vocabulary, that is certain rhythmical motives reoccur several times in a given piece. The rhythmic structure depends mostly upon the musical genre and composer. It seems to be rather difficult to devise a general prior model that would work well in a large spectrum of styles. Nevertheless, for a given genre, we expect a simple prior to capture enough structure sufficient for good transcription. For example, for the Beatles dataset, we have estimated the prior by counting from the original score of "Yesterday". The statistics are fairly close to that of "Michelle". The good results on the test set can be partially accounted for the fact that both pieces have a similar rhythmical structure.

Conditioned on the score, the tempo tracking model is a linear dynamical system. We have optimized several tempo models using EM where we have varied the dimension of tempo variables $z$. The test results suggest that increasing the dimensionality of $z$ improves the likelihood. However, increase in the likelihood of the whole dataset does not translate directly to overall better quantization results (as measured by edit distance). We observe that models trained on the whole training data fail consistently for some subjects, especially professional classical pianists. Perhaps interestingly, if we train "custom" models specifically optimized for the same subjects, we can improve results significantly also on test cases. This observation suggests a kind of multimodality in the parameter space where modes correspond to different performer regimes. It seems that a Kalman filter is able to capture the structure in expressive timing deviations. However, when averaged over all subjects, these details tend to be wiped out, as suggested by the quantization results that do not vary significantly among models of different dimensions.

A related problem with the edit distance measure is that under an "average" model, the likelihood of the desired score (e.g., original score of "Michelle") may have a lower likelihood than a solution found by an inference method. In such cases increasing the likelihood may even decrease the edit distance. In some test cases we even observe solutions with a higher likelihood than the original notation where all notes are wrong. In most of these cases, the tempo trajectory of the solution correspond to the half or twice of the original tempo so consequently all note durations are halved or doubled (e.g., all whole notes are notated as half notes, all half notes as quarters e.t.c.). Considering the fact that the model is "self initializing" its tempo, that is we assume a broad uncertainty a-priori, the results are still satisfactory from a practical application perspective.

One potential shortcoming of our model is that it takes only timing information of onsets into account. In reality, we believe that pitch and melodic grouping as well as articulation (duration between note onsets and offsets) and dynamics (louder or softer) provide useful additional information for tempo tracking as well as quantization. Moreover, current model assumes that all onsets are equally relevant for estimation. That is probably in general not true: for example, a kick-drum should provide more information about the tempo than a flute. On the other hand, our simulations suggest that even from such a limited model one can obtain quite satisfactory results, at least for simple piano music.

It is somewhat surprising, that SMC, basically a method that samples from the filtering distribution outperforms an MCMC method such as SA that is specifically designed for finding the MAP solution given all observations. An intuitive explanation for relatively poorer MCMC results is that MCMC proceeds first by proposing a global solution and then tries to improve it by local adjustments. A human transcriber, on the other hand, would listen to shorter segments of music and gradually write down the score. In that respect, the sequential update schema of SMC seems to be more natural for the rhythm transcription problem. Similar results, where SMC outperforms MCMC are already reported in the literature, e.g., in the so-called "Growth Monte Carlo" for generating self-avoiding random walks (Liu, Chen, & Logvinenko, 2001). It seems that for a large class of dynamical problems, including rhythm transcription, sequential updating is preferable over batch methods.

We note that theoretical convergence results for SA require the use of a logarithmic cooling schedule. It seems that our cooling schedule was too fast to meet this requirement; so one has to be still careful in interpreting the poor performance as a negative SA result. We maintain that by using a richer neighborhood structure in the configuration space (e.g., by using a block proposal distribution) and a slower cooling schedule, SA results can be improved significantly. Moreover, MCMC methods can be also be modified to operate sequentially, for example see (Marthi, Pasula, Russell, & Peres, 2002).

Another family of inference methods for switching state space models rely on deterministic approximate methods. This family includes variational approximations (Ghahramani & Hinton, 1998) and expectation propagation (Heskes, 2002). It remains an interesting open question whether deterministic approximation methods provide an advantage in terms of computation time and accuracy; in particular for the quantization problem and for other switching state space models. A potential application of the deterministic approximation techniques in a MCMC schema can be in designing proposal distributions that extend over several time slices. Such a schema would circumvent the burden for computing the optimal proposal distribution exhaustively hence allowing more global moves for the sampler.

Our current results suggest the superiority of SMC for our problem. Perhaps the most important advantage of SMC is that it is essentially an "anytime" algorithm; if we have a faster computer we can increase the number of particles to make use of the additional computational power. When computing time becomes short one can decrease the number of samples. These features make SMC very attractive for real-time applications where one can easily tune the quality/computation-time tradeoff.

Motivated by the practical advantages of SMC and our positive simulation results, we have implemented a prototype of SMC method in real-time. Our current computer system (a 800 MHz P3 laptop PC running MS Windows) allows us to use up to 5 particles with almost no delay even during busy passages. We expect to significantly improve the efficiency by translating the MATLAB© constructs to native C code. Hence, the method can be used as a tempo tracker in an automatic interactive performance system and as a quantizer in an automatic score typesetting program.

## Acknowledgments

This research is supported by the Technology Foundation STW, applied science division of NWO and the technology programme of the Dutch Ministry of Economic Affairs. We would like to thank the associate editor Daphne Koller and the anonymous reviewers for their comments that helped us significantly to improve the article. We would also like to thank to Ric Ashley, Peter Desain, Henkjan Honing and Paul Trilsbeek for their suggestions and contributions in data collection. Moreover we gratefully acknowledge the pianists from Northwestern University and Nijmegen University for their excellent performances.

## Appendix A. A generic prior model for quantization locations $c$

In traditional western music notation, note durations are generated by recursive subdivisions starting from a whole note, hence it is also convenient to generate quantization locations in a similar fashion by regular subdivisions. We decompose a quantization location into an integer part and a fraction: $c = \lfloor c \rfloor + (c \mod 1)$. For defining a prior, we will only use the fraction.

The set of all fractions can be generated by recursively subdividing the unit interval $[0, 1)$. We let $\mathcal{S} = [s_i]$ denote a subdivision schema, where $[s_i]$ is a (finite) sequence of arbitrary integers (usually small primes such as 2,3 or 5). The choice of a particular $\mathcal{S}$ depends mainly on the assumed time signature. We generate the set of fractions $C$ as follows: At first iteration, we divide the unit interval into $s_1$ intervals of equal length and append the endpoints $c'$ of resulting intervals into the set $C$. At each following iteration $i$, we subdivide all intervals generated by the previous iteration into $s_i$ equal parts and append all resulting endpoints to $C$. Note that this procedure generates a regular grid where two neighboring grid points have the distance $1/\prod_i s_i$. We denote the iteration number at which the endpoint $c'$ is first inserted to $C$ as the *depth* of $c'$ (with respect to $\mathcal{S}$). This number will be denoted as $d(c'|\mathcal{S})$. It is easy to see that this definition of $d$ coincides with the number of significant bits to represent $c \mod 1$ when $\mathcal{S} = [2,\ 2, \dots]$.

As an illustrative example consider the subdivision $\mathcal{S} = [3, 2]$. At the first iteration, the unit interval is divided into $s_1 = 3$ equal intervals, and the resulting endpoints 0, 1/3, and 2/3 are inserted into $C$ with depths $d(0) = d(1/3) = d(2/3) = 1$. At the second iteration, the new endpoints 1/6, 3/6 and 5/6 are inserted to $C$ and are assigned the depth 2.

Given an $\mathcal{S}$, we can define a distribution on quantization locations

$$p(c_k|\mathcal{S}) \propto \exp(-\lambda d(c_k \mod 1|\mathcal{S}))$$

If we wish to consider several time signatures, i.e., different subdivision schemata, we can interpret $\mathcal{S}$ as a hidden indicator variable and define a prior $p(\mathcal{S})$. In this case, the prior becomes a multinomial mixture given by $p(c_k) = \sum_{\mathcal{S}} p(c_k|\mathcal{S})p(\mathcal{S})$. For further details and empirical results justifying such a choice see (Cemgil et al., 2000).

## Appendix B. Derivation of two pass Kalman filtering Equations

Consider a Gaussian potential with mean $\mu$ and covariance $\Sigma$ defined on some domain indexed by $x$.

$$\phi(x) \quad = \quad Z \times \mathcal{N}(\mu, \Sigma) = Z|2\pi\Sigma|^{-\frac{1}{2}} \exp(-\frac{1}{2}(x-\mu)^T \Sigma^{-1}(x-\mu)) \tag{28}$$

where $\int dx \phi(x) = Z > 0$. If $Z = 1$ the potential is normalized. The exponent in Eq. 28 is a quadratic form so the potential can be written as

$$\phi(x) \quad = \quad \exp(g + h^T x - \frac{1}{2} x^T K x) \tag{29}$$

where

$$K = \Sigma^{-1} \qquad h = \Sigma^{-1}\mu \qquad g = \log Z + \frac{1}{2}\log|\frac{K}{2\pi}| - \frac{1}{2}h^T K^{-1} h$$

To denote a potential in canonical form we will use the notation

$$\phi(x) \quad = \quad Z \times \mathcal{N}(\mu, \Sigma) \equiv [h, K, g]$$

and we will refer to $g$, $h$ and $K$ as *canonical* parameters. Now we consider a Gaussian potential on $(x_1, x_2)^T$. The canonical representation is

$$\phi(x_1, x_2) \quad = \quad \left[ \begin{pmatrix} h_1 \\ h_2 \end{pmatrix}, \begin{pmatrix} K_{11} & K_{12} \\ K_{21} & K_{22} \end{pmatrix}, g \right]$$

In models where several variables are interacting, one can find desired quantities by applying three basic operations defined on Gaussian potentials. Those are *multiplication*, *conditioning*, and *marginalization*. The multiplication of two Gaussian potentials on the same index set $x$ follows directly from Eq. 29 and is given by

$$\begin{aligned} \phi'(x) &= \phi_a(x) \times \phi_b(x) \\ [h', K', g'] &= [h_a, K_a, g_a] \times [h_b, K_b, g_b] = [h_a + h_b, K_a + K_b, g_a + g_b] \end{aligned}$$

If the domain of $\phi_a$ and $\phi_b$ only overlaps on a subset, then potentials are extended to the appropriate domain by appending zeros to the corresponding dimensions.

The marginalization operation is given by

$$\phi(x_1) = \int_{x_2} \phi(x_1, x_2) = [h_1 - K_{12}K_{22}^{-1}h_2, K_{11} - K_{12}K_{22}^{-1}K_{21}, g']$$

where $g' = g - \frac{1}{2}\log|K_{22}/2\pi| + \frac{1}{2}{h_2}^T(K_{22})^{-1}h_2$ and $g$ is the initial constant term of $\phi(x_1, x_2)$. The conditioning operation is given by

$$\phi(x_1, x_2 = \hat{x}_2) = [h_1 - K_{12}\hat{x}_2, K_{11}, g']$$

where $g' = g + h_2^T\hat{x}_2 - \frac{1}{2}\hat{x}_2^T K_{22}\hat{x}_2$.

### B.1 The Kalman Filter Recursions

Suppose we are given the following linear model subject to noise

$$\begin{aligned} z_k &= Az_{k-1} + \zeta_k \\ y_k &= Cz_k + \epsilon_k \end{aligned}$$

where $A$ and $C$ are constant matrices, $\zeta_k \sim \mathcal{N}(0, Q)$ and $\epsilon_k \sim \mathcal{N}(0, R)$

The model encodes the joint distribution

$$p(z_{1:K}, y_{1:K}) = \prod_{k=1}^{K} p(y_k|z_k)p(z_k|z_{k-1}) \quad (30)$$

$$p(z_1|z_0) = p(z_1) \quad (31)$$

$$\begin{aligned} p(z_1) &= [P^{-1}\mu, P^{-1}, -\frac{1}{2}\log|2\pi P| - \frac{1}{2}\mu^T P^{-1}\mu] \\ p(y_1|z_1) &= \left[\begin{pmatrix} 0 \\ 0 \end{pmatrix}, \begin{pmatrix} C^T R^{-1} C & -C^T R^{-1} \\ -R^{-1}C & R^{-1} \end{pmatrix}, -\frac{1}{2}\log|2\pi R|\right] \\ p(y_1 = \hat{y}_1|z_1) &= [0 + C^T R^{-1}\hat{y}_1, C^T R^{-1} C, -\frac{1}{2}\log|2\pi R| - \frac{1}{2}\hat{y}_1^T R^{-1}\hat{y}_1] \\ p(z_2|z_1) &= \left[\begin{pmatrix} 0 \\ 0 \end{pmatrix}, \begin{pmatrix} A^T Q^{-1} A & -A^T Q^{-1} \\ -Q^{-1}A & Q^{-1} \end{pmatrix}, -\frac{1}{2}\log|2\pi Q|\right] \\ &\dots \end{aligned}$$

#### B.1.1 Forward Message Passing

Suppose we wish to compute the likelihood

$$p(y_{1:K}) = \int_{z_K} p(y_K|z_K) \dots \int_{z_2} p(z_3|z_2)p(y_2|z_2) \int_{z_1} p(z_2|z_1)p(y_1|z_1)p(z_1)$$

[7]We can compute this integral by starting from $z_1$ and proceeding to $z_K$. We define forward "messages" $\alpha$ as

7. We let $\int_z \equiv \int dz$

- $\alpha_{1|0} = p(z_1)$
- $k = 1 : K$
  - $\alpha_{k|k} = p(y_k = \hat{y}_k|z_k)\alpha_{k|k-1}$
  - $\alpha_{k+1|k} = \int_{z_k} p(z_{k+1}|z_k)\alpha_{k|k}$

The forward recursion is given by

- $\alpha_{1|0} = [P^{-1}\mu, P^{-1}, -\frac{1}{2}\log|2\pi P| - \frac{1}{2}\mu^T P^{-1}\mu]$
- $k = 1 \dots K$
  - $\alpha_{k|k} = [h_{k|k}, K_{k|k}, g_{k|k}]$
    $h_{k|k} = C^T R^{-1}\hat{y}_k + h_{k|k-1}$
    $K_{k|k} = C^T R^{-1} C + K_{k|k-1}$
    $g_{k|k} = g_{k|k-1} - \frac{1}{2}\log|2\pi R| - \frac{1}{2}\hat{y}_1^T R^{-1}\hat{y}_k$
  - $\alpha_{k+1|k} = [h_{k+1|k}, K_{k+1|k}, g_{k+1|k}]$
    $M_k = (A^T Q^{-1} A + K_{k|k})^{-1}$
    $h_{k+1|k} = Q^{-1} A M_k h_{k|k}$
    $K_{k+1|k} = Q^{-1} - Q^{-1} A M_k A^T Q^{-1}$
    $g_{k+1|k} = g_{k|k} - \frac{1}{2}\log|2\pi Q| + \frac{1}{2}\log|2\pi M_k| + \frac{1}{2}h_{k|k}^T M_k h_{k|k}$

### B.1.2 Backward Message Passing

We can compute the likelihood also by starting from $y_K$.

$$p(y_{1:K}) \quad = \quad \int_{z_1} p(z_1)p(y_1|z_1)\int_{z_2} p(z_2|z_1)p(y_2|z_2)\dots\int_{z_K} p(z_K|z_{K-1})p(y_K|z_K)$$

In this case the backward propagation can be summarized as

- $\beta_{K|K+1} = 1$
- $k = K \dots 1$
  - $\beta_{k|k} = p(y_k = \hat{y}_k|z_k)\beta_{k|k+1}$
  - $\beta_{k-1|k} = \int_{z_k} p(z_k|z_{k-1})\beta_{k|k}$

The recursion is given by

- $[h^*_{K|K+1}, K^*_{K|K+1}, g^*_{K|K+1}] = [0, 0, 0]$
- $k = K \dots 1$
  - $\beta_{k|k} = [h^*_{k|k}, K^*_{k|k}, g^*_{k|k}]$
    $h^*_{k|k} = C^T R^{-1}\hat{y}_k + h^*_{k|k+1}$
    $K^*_{k|k} = C^T R^{-1} C + K^*_{k|k+1}$

$$g^*_{k|k} = -\tfrac{1}{2}\log|2\pi R| - \tfrac{1}{2}\hat{y}_k^T R^{-1}\hat{y}_k + g^*_{k|k+1}$$

$- \;\beta_{k-1|k} = [h^*_{k-1|k}, K^*_{k-1|k}, g^*_{k-1|k}]$

$$M^*_k = (Q^{-1} + K^*_{k|k})^{-1}$$
$$h^*_{k-1|k} = A^T Q^{-1} M^*_k h^*_{k|k}$$
$$K^*_{k-1|k} = A^T Q^{-1}(Q - M^*_k)Q^{-1}A$$
$$g^*_{k-1|k} = g^*_{k|k} - \tfrac{1}{2}\log|2\pi Q| + \tfrac{1}{2}\log|2\pi M^*_k| + \tfrac{1}{2}h^{*T}_{k|k} M^*_k h^*_{k|k}$$

### B.2 Kalman Smoothing

Suppose we wish to find the distribution of a particular $z_k$ given all the observations $y_{1:K}$. We just have to combine forward and backward messages as

$$\begin{aligned} p(z_k|y_{1:K}) &\propto p(y_{k+1:K}, z_k, y_{1:k}) \\ &= p(y_{1:k}, z_k)p(y_{k+1:K}|z_k) \\ &= \alpha_{k|k} \times \beta_{k|k+1} \\ &= [h_{k|k} + h^*_{k|k+1}, K_{k|k} + K^*_{k|k+1}, g_{k|k} + g^*_{k|k+1}] \end{aligned}$$

## Appendix C. Rao-Blackwellized SMC for the Switching State space Model

We let $i = 1 \ldots N$ be an index over particles and $s = 1 \ldots S$ an index over states of $\gamma$. We denote the (unnormalized) filtering distribution at time $k-1$ by

$$\phi^{(i)}_{k-1} \;\hat{=}\; p(y_{0:k-1}, z_{k-1}|\gamma^{(i)}_{1:k-1})$$

Since $y_{0:k-1}$ are observed, $\phi^{(i)}_{k-1}$ is a Gaussian potential on $z_{k-1}$ with parameters $Z^{(i)}_{k-1} \times \mathcal{N}(\mu^{(i)}_{k-1}, \Sigma^{(i)}_{k-1})$. Note that the normalization constant $Z^{(i)}_{k-1}$ is the data likelihood $p(y_{0:k-1}|\gamma^{(i)}_{1:k-1}) = \int dz_k \phi^{(i)}_{k-1}$. Similarly, we denote the filtered distribution at the next slice conditioned on $\gamma_k = s$ by

$$\begin{aligned} \phi^{(s|i)}_k &\;\hat{=}\; \int dz_{k-1} p(y_k|z_k)p(z_k|z_{k-1}, \gamma_k = s)\phi^{(i)}_{k-1} \\ &= p(y_{0:k}, z_k|\gamma^{(i)}_{1:k-1}, \gamma_k = s) \end{aligned} \tag{32}$$

We denote the normalization constant of $\phi^{(s|i)}_k$ by $Z^{(s|i)}_k$. Hence the joint proposal on $s$ and $(i)$ is given by

$$\begin{aligned} q^{(s|i)}_k &= \int dz_k \phi^{(s|i)}_k \times p(\gamma_k = s, \gamma^{(i)}_{1:k-1}) \\ &= p(\gamma_k = s, \gamma^{(i)}_{1:k-1}, y_{0:k}) \end{aligned}$$

The outline of the algorithm is given below:

- Initialize. For $i = 1 \ldots N$, $\phi^{(i)}_0 \leftarrow p(y_0, x_0)$

- For $k = 1 \dots K$
  - For $i = 1 \dots N$, $s = 1 \dots S$

    Compute $\phi_k^{(s|i)}$ from $\phi_{k-1}^{(i)}$ using Eq.32.

    $q_k^{(s|i)} \leftarrow Z_k^{(s|i)} \times p(\gamma_k = s, \gamma_{1:k-1}^{(i)})$
  - For $i = 1 \dots N$

    Select a tuple $(s|j) \sim q_k$

    $\gamma_{1:k}^{(i)} \leftarrow (\gamma_{1:k-1}^{(j)}, \gamma_k = s)$

    $\phi_k^{(i)} \leftarrow \phi_k^{(s|j)}$

    $w_k^{(i)} \leftarrow \sum_s q_k^{(s|j)}$

Note that the procedure has a "built-in" resampling schema for eliminating particles with small importance weight. Sampling jointly on $(s|i)$ is equivalent to sampling a single $s$ for each $i$ and then resampling $i$ according to the weights $w_k^{(i)}$. One can also check that, since we are using the optimal proposal distribution of Eq.27, the weight at each step is given by $w_k^{(i)} = p(\gamma_{1:k-1}^{(i)}, y_{0:k})$.